%% file: main_cvpr2024.tex
\documentclass[10pt,twocolumn,letterpaper]{article}

\usepackage[pagenumbers]{cvpr} 

\usepackage[dvipsnames]{xcolor}
\definecolor{citecolor}{HTML}{0071bc}
\definecolor{linkcolor}{RGB}{215,   0,   64}
\usepackage[pagebackref,breaklinks=true,letterpaper=true,colorlinks=true,linkcolor=linkcolor,citecolor=citecolor,bookmarks=false]{hyperref}

\usepackage{graphicx}
\usepackage{amsmath}
\usepackage{url}            
\usepackage{booktabs}       
\usepackage{amsfonts}       
\usepackage{nicefrac}       
\usepackage{microtype}      
\usepackage{algorithm}
\usepackage{comment}
\usepackage{pifont}
\usepackage[noend]{algpseudocode}
\usepackage{bbm}
\usepackage{dsfont}
\usepackage{easylist}
\usepackage{xspace}
\usepackage{wrapfig}
\usepackage{multirow} 
\usepackage{amssymb}
\usepackage{caption}
\usepackage{subcaption}
\usepackage{float}
\usepackage{xcolor,colortbl}
\usepackage[normalem]{ulem}

\definecolor{skyblue1}{RGB}{114, 159, 207}
\definecolor{myred}{RGB}{187,   0,   0}
\definecolor{mygreen}{RGB}{20,   132,   0}
\definecolor{mytangoblue}{RGB}{59, 109, 172}
\definecolor{mytangoorange}{RGB}{243, 121, 33}
\definecolor{mypurple}{RGB}{150,   20,   150}
\definecolor{mygrey}{RGB}{150,   150,   150}
\definecolor{mycell1}{RGB}{240,   230,   230}
\definecolor{mycell2}{RGB}{240,   230,   230}

\newcommand{\myparagraph}[1]{\vspace{0pt}\noindent{\bf #1}}
\newcommand{\tablestyle}[2]{\setlength{\tabcolsep}{#1}\renewcommand{\arraystretch}{#2}\centering\footnotesize} 

\newlength\savewidth\newcommand\shline{\noalign{\global\savewidth\arrayrulewidth
  \global\arrayrulewidth 1pt}\hline\noalign{\global\arrayrulewidth\savewidth}}
\newcommand\midline{\noalign{\global\savewidth\arrayrulewidth
  \global\arrayrulewidth 0.2pt}\hline\noalign{\global\arrayrulewidth\savewidth}}
  
\newcommand{\ours}{\textbf{\scshape{decola}}\xspace}
\newcommand{\oursbig}{\textbf{\scshape{DECOLA}}\xspace}

\newcommand{\lbltbl}[1]{\label{tbl:#1}}
\newcommand{\base}{baseline\xspace}
\newcommand{\goodnum}[2]{#1 \textcolor{mygreen}
{\scriptsize{(+#2)}}}
\newcommand{\gooddelta}[1]{\textcolor{mygreen}
{\scriptsize{(+#1)}}}
\newcommand{\badnum}[2]{#1 \textcolor{myred}
{\scriptsize{(-#2)}}}

\definecolor{chosencolor}{gray}{.8}

\def\paperID{1025} 
\def\confName{CVPR}
\def\confYear{2024}

\title{Language-conditioned Detection Transformer}

\author{Jang Hyun Cho \\
UT Austin \\
{\tt\small janghyuncho7@utexas.edu}
\and
Philipp Kr\"ahenb\"uhl \\
 UT Austin \\
{\tt\small philkr@cs.utexas.edu}
}

\begin{document}

\maketitle

\begin{abstract}
We present a new open-vocabulary detection framework. 
Our framework uses both image-level labels and detailed detection annotations when available.
Our framework proceeds in three steps. 
We first train a language-conditioned object detector on fully-supervised detection data. 
This detector gets to see the presence or absence of ground truth classes during training, and conditions prediction on the set of present classes. 
We use this detector to pseudo-label images with image-level labels. 
Our detector provides much more accurate pseudo-labels than prior approaches with its conditioning mechanism. 
Finally, we train an unconditioned open-vocabulary detector on the pseudo-annotated images. 
The resulting detector, named \ours, shows strong zero-shot performance in open-vocabulary LVIS benchmark as well as direct zero-shot transfer benchmarks on LVIS, COCO, Object365, and OpenImages. 
\ours outperforms the prior arts by \textbf{17.1} AP$_\text{rare}$ and \textbf{9.4} mAP on zero-shot LVIS benchmark.
\ours achieves state-of-the-art results in various model sizes, architectures, and datasets by only training on open-sourced data and academic-scale computing. 
Code is available at \url{https://github.com/janghyuncho/DECOLA}.

\end{abstract}

\vspace{-4mm}
\section{Introduction}

\begin{figure}[t]
    \centering
    \leftskip2mm
    \includegraphics[width=1.01\linewidth]{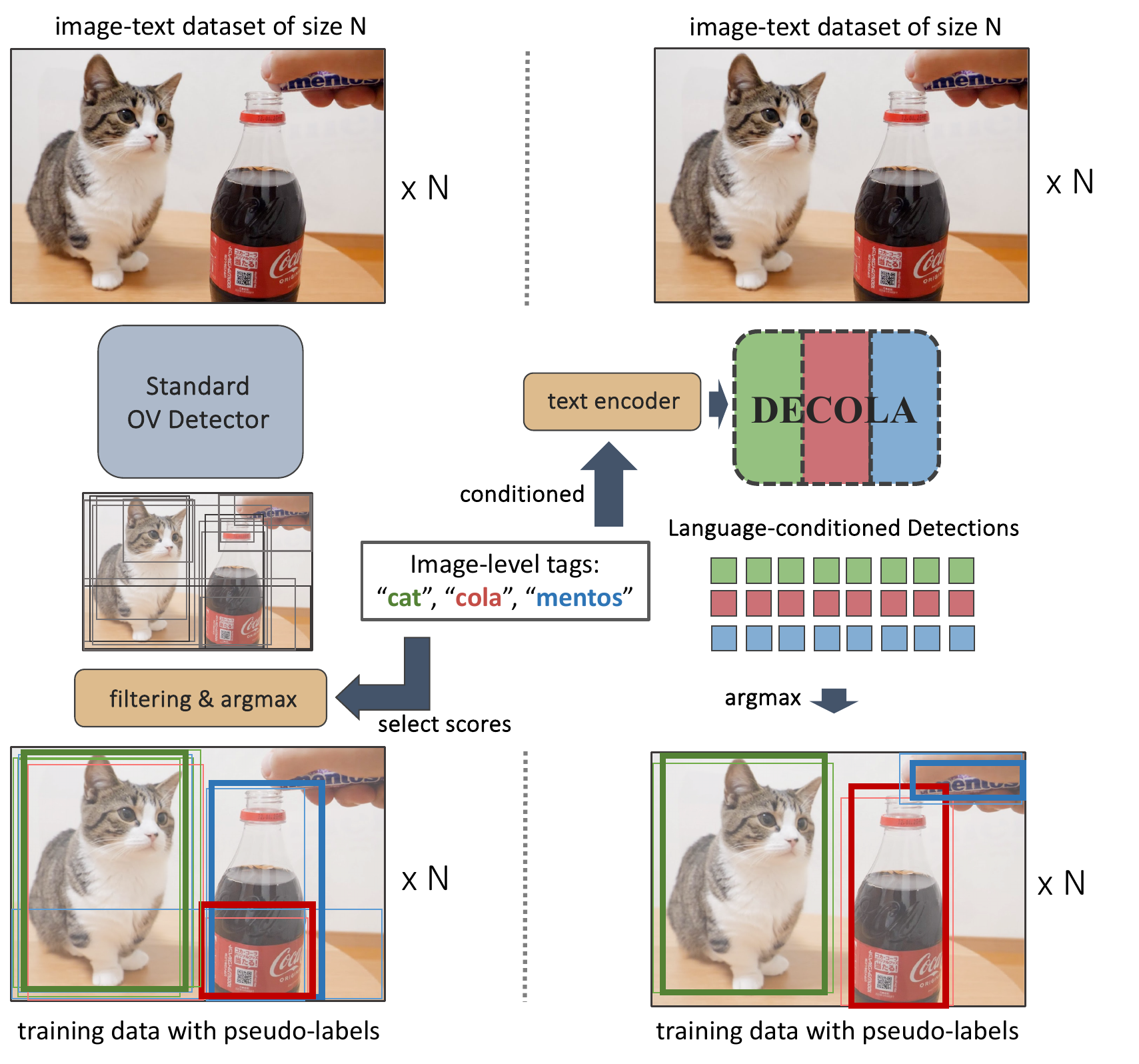}
    \vspace{-1mm}
    \begin{subfigure}{0.45\columnwidth}
        \captionsetup{
        width=\linewidth}
        \caption{Standard self-training}
        \label{fig:standard_ov_st}
    \end{subfigure}
    \quad\;
    \begin{subfigure}{0.45\columnwidth}
        \captionsetup{width=0.75\linewidth}       
        \caption{\ours self-training}
        \label{fig:decola_st}
    \end{subfigure}
    \caption{An illustration of how standard open-vocabulary detectors and \ours generate pseudo-labels using image-level data. 
    Standard detectors use image-level information later in the pipeline after initial box proposals, which may result in low coverage of unseen classes (\emph{e.g.,} ``\textbf{\textcolor{mytangoblue}{mentos}}'' and ``\textbf{\textcolor{myred}{cola}}'').
    \ours adjusts the prediction to the information and ensures sufficient coverage.
    }
    \vspace{-7mm}
    \label{fig:teaser}
\end{figure}

Object detection has seen immense progress over the past decade.
Classical object detectors reason over datasets of fixed predefined classes.
This simplifies the design, training, and evaluation of new methods, and allows for rapid prototyping~\cite{rcnn,fastrcnn,cai2018cascade,htc,zhou2021probabilistic,detr,zhu2020deformable,zhang2022dino,conditional_detr,dab_detr,hdetr,anchor_detr,group_detr}.
However, it complicates deployment to downstream applications too.
A classical detector requires a new dataset to further finetune for every new concept it encounters.
Collecting sufficient data for every new concept is not scalable~\cite{gupta2019lvis}. 
Open-vocabulary detection offers an alternative~\cite{vild,minderer2022simple,detic,ovr-cnn,wu2023baron,ov-detr}. 
Open-vocabulary detectors reason about any arbitrary concept with free-form text, using the generalization ability of vision-language models.
Yet, common open-vocabulary detectors reuse classical detectors and either replace the last classification layer with~\cite{detic,ovr-cnn,vild,minderer2022simple}, or fuse box feature with~\cite{ov-detr,fvlm} text representation from pretrained vision-language model. 
The inner workings of the detector remain unchanged.

In this paper, we introduce a transformer-based object detector that adjusts its inner workings to any arbitrary set of concepts represented in language. 
The detector considers only the queried set of concepts as foreground and disregards any other objects as background.  
It learns to adapt detection to the language embedding of queried concepts at run-time. 
Specifically, the detector conditions proposal generation with respect to the text embedding of each queried concept and refines the conditioned proposals into predictions. 
Our \underline{\textit{detection transformer conditioned on language}} (\ours) offers a powerful alternative to classical architectures in open-vocabulary detection. 
Adapting the detector to language leads to stronger generalization to unseen concepts, and largely enhances self-training on weakly-labeled data. 

\ours's ability to readily adapt to any queried concepts makes it particularly suitable for pseudo-labeling weakly-labeled data. 
Internet data, specifically images with paired text, is highly abundant and semantically rich~\cite{laion,sharma2018conceptual,ILSVRC15}.
The best vision models today build on this massive amount of weakly labeled data~\cite{clip,openclip,glip,blip,eva,lit,vit22b}.
\ours leverages the same data to produce high-quality object detection labels from image-level annotations alone.  
\ours takes the image-level tags or text descriptions from the weakly labeled data and generates conditioned predictions as pseudo-labels. 
It efficiently processes multiple texts in parallel and only adds minimal computational overhead compared to the standard pseudo-labeling process.
We finetune \ours on this rich detection dataset of pseudo-annotations and achieve the state-of-the-art open-vocabulary detector.

We evaluate our detector on popular open-vocabulary detection benchmarks on the LVIS dataset~\cite{gupta2019lvis,vild,ovr-cnn}. 
The final model improves the state-of-the-art methods by \textbf{4.4} and \textbf{4.9} AP$_{\text{novel}}$ on open-vocabulary LVIS~\cite{vild} benchmark, and \textbf{5.9} and \textbf{5.4} AP$_{\text{rare}}$ on standard LVIS~\cite{gupta2019lvis} benchmark, with ResNet-50~\cite{he2016deep} and Swin-B~\cite{liu2021Swin} backbones, respectively. 
Our largest model with Swin-L achieves \textbf{10.4} AP$_{\text{rare}}$ and \textbf{3.6} mAP improvement.
Furthermore, \ours largely outperforms the state-of-the-art for \textit{direct zero-shot transfer} benchmark on LVIS, by \textbf{12.0}  and \textbf{17.1} AP$_{\text{rare}}$ on LVIS minival and LVIS v1.0, respectively. 
\ours consistently improves frequent, common, and rare classes altogether for different backbones and detection frameworks.
Much of this improvement is driven by stronger pseudo-labeling capabilities.
All our models are trained using open-sourced datasets with academic-scale computing. 
We open-source our code, pseudo-annotations, and checkpoints of all the model scales.

\section{Related Work}

\begin{figure*}[h]
    \centering
    \includegraphics[width=0.9\linewidth]{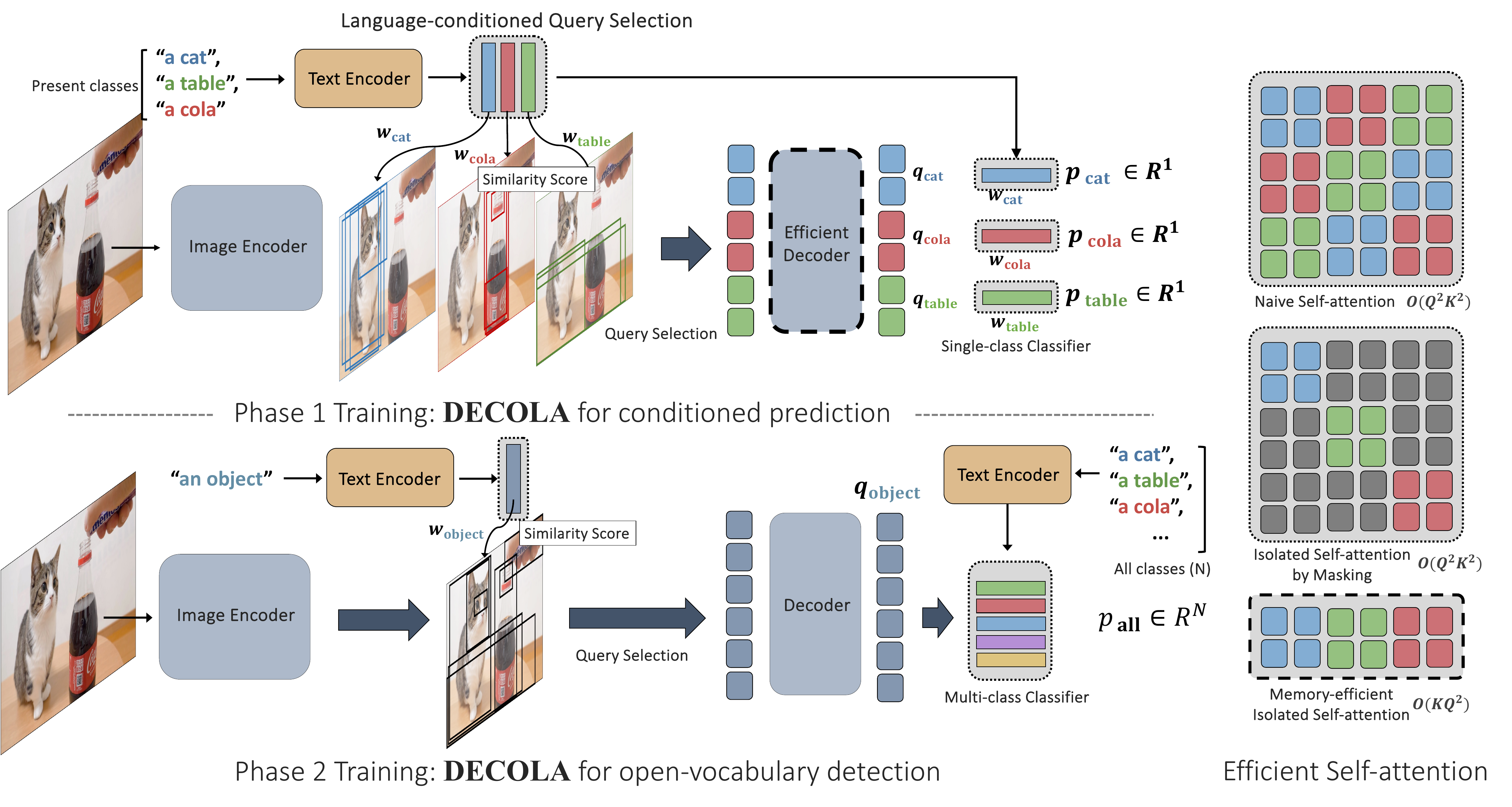}
    \vspace{-4mm}
    \caption{Overview of \ours Phase 1 for conditioned prediction (\textbf{top}) and Phase 2 for open-vocabulary detection (\textbf{bottom}). 
    For language-conditioned detection, each text embedding directly parameterizes the objectness function for each class. 
    In Phase 2, the language-condition reads ``\texttt{an object}'' instead of particular class names, and predicts multi-class scores over all classes after decoding layers. 
    }
    \vspace{-2mm}
    \vspace{-3mm}
    \label{fig:pipeline}
\end{figure*}

\myparagraph{Open-vocabulary detection}
aims to detect objects of categories beyond the vocabulary of the training classes.
A common solution is to inject language embeddings of class names in the last classification layer. 
OVR-CNN~\cite{ovr-cnn} pretrains a detector on image-caption data using BERT model~\cite{bert} as language embedding. 
ViLD~\cite{vild} trains a detector with CLIP text encoder~\cite{clip} as language embedding with additional knowledge distillation~\cite{kd} between predicted box features and the image encoder of CLIP. 
Detic~\cite{detic} improves the above approaches through weakly-supervised learning on image-level annotations.
RegionCLIP~\cite{zhong2022regionclip} introduces an intermediate pretraining step to better align CLIP to box features. 
BARON~\cite{wu2023baron} improves the alignment between text and image encoders by extracting \textit{bag of regions} as additional supervision. 
F-VLM~\cite{fvlm} simplifies the training pipeline of open-vocabulary detection and explores the limit of the frozen vision-language model.
All of the models above take the design of the object detector as granted, and inject language in the last classification layer of the network.
We take a different approach and design a detector that adapts predictions to particular categories of interest. 

\myparagraph{Open-vocabulary DETR}
integrates DETR architecture into open-vocabulary detection.
OWL~\cite{minderer2022simple} introduces a simple ViT architecture using pretrained CLIP and finetune with the DETR objective.
OWLv2 uses self-training to further improve the performance~\cite{owlv2}. 
OV-DETR~\cite{ov-detr} fuses features of a pretrained CLIP model with DETR object queries. 
Architecturally, OV-DETR is closest to \ours. 
Both OV-DETR and \ours condition predictions on the text representation of classes.
OV-DETR uses the original DETR queries and expands each with a CLIP feature per class for all classes. 
This leads to a quadratic number of queries, growing with the original DETR queries and with the classes considered. 
On the other hand, \ours explicitly controls the first-stage predictions (proposals) by formulating the scoring function to respect to the text embedding of each queried class at run-time. 
We visualize this difference in Figure~\ref{fig:diff} in the supplementary.
The advantage is that we entirely remove inter-class competition and process a manageable amount of queries each focusing on a specific class, and running as fast as the vanilla Deformable DETR.
This ability to freely adjust inner workings deviates \ours from prior works; {it expands detection data through high-quality pseudo-labeling and achieves state-of-the-art results.}

\myparagraph{Large-vocabulary object detection}
shares similar goals with open-vocabulary detection.
Both learn from naturally long-tail data over large vocabularies.
Vanilla large-vocabulary detectors are often ill-calibrated: The detector's final classification layer favors frequently seen objects over rare ones. 
This imbalance is usually addressed through a change in loss~\cite{seesaw,eql,eqlv2,ecm,zhou2021probabilistic}, or leveraging additional weakly labeled data for self-training~\cite{copypaste,mosaic,selftrain,pd}.
In large-vocabulary detection, R-CNN-based frameworks~\cite{rcnn,fastrcnn,cai2018cascade,htc,zhou2021probabilistic} dominate despite DETR-style architectures~\cite{ouyang2022nms,zhang2022dino,hdetr,group_detr} having long surpassed them on standard benchmarks~\cite{coco}. 
DETR automatically assigns object queries to output classes, and thus it learns to more heavily focus queries on common classes.
We show that language-conditioning helps address this calibration issue.
Specifically, it removes \emph{inter-class competition} in the training objective as queries are no longer shared across categories.
As a result, \ours equally focuses on as many rare classes as frequent ones whenever they are present in an image.
This yields a DETR-style detector that is competitive with the best R-CNN-based large-vocabulary detectors.

\section{Preliminaries}
\label{prelim_detr}
Detection transformers (DETR)~\cite{detr} build an object detection pipeline as a single feed-forward network.
The network transforms object queries, arbitrary feature vectors, into labeled bounding boxes through a series of cross-attention layers in a decoder architecture.
Vanilla DETR~\cite{detr} learns object queries as free-form parameters, while modern DETRs architectures~\cite{zhu2020deformable,zhang2022dino,hdetr,dab_detr,anchor_detr} adopt a two-stage paradigm similar to RCNNs~\cite{ren2015faster}.
This query mechanism controls much of the inner workings of the detector.
Queries determine what image regions the detector focuses on, and what object classes are prioritized.

\myparagraph{Query selection.} Modern DETR architectures use image-dependent query selection, analogous to R-CNN's proposal generation~\cite{ren2015faster}.
An objectness function $s$ scores each grid location $(i,j)$ in the image using a feature  $x_{i,j}$ extracted from the transformer encoder.
The top-$k$ scored regions proceed to the second stage as object queries $Q$: 
\begin{equation}
s(x_{i,j})= \langle x_{i,j}, w \rangle, \quad Q=\texttt{topk}_{x_{i,j}}(s(x_{i,j}))\label{classic_obj}.
\end{equation}
Here, $w$ are the parameters of the objectness predictor.
Each selected query produces a series of predictions that are refined over multiple iterations, similar to Cascade R-CNN~\cite{cai2018cascade}. 
The final prediction $\vec{p}$ contains scores over all classes $C$ and an associated box. 
At a high level, DETR and R-CNNs share the same motivation: first, localize \textit{all} objects in a scene, then refine their predictions. 

\myparagraph{Training objective.} During training, DETR assigns each object query to an object or marks it as background. 
This allows DETR to learn non-overlapping object queries without post-processing such as non-maximum-suppression. 
The Hungarian matching algorithm finds the optimal assignment between all predictions $P$ and all ground truth $G$, minimizing the loss function as matching cost: 
\begin{equation}
\sigma^* = \arg\min_{\sigma\in \mathfrak{S}} \ell(P,G|\sigma)\label{detr_objective}
\end{equation}
where $\mathfrak{S}$ captures all possible assignments from $P$ to $G$.
For each assigned prediction, the loss $\ell$ maximizes its class log-likelihood and fits its bounding box. 
For unassigned predictions, the loss $\ell$ reduces both the objectness score $s$ and class log-likelihood for all classes.

\section{\oursbig}
\label{sec:method}

Our detection transformer conditioned on language, \ours, changes the DETR architecture in one remarkable way:
Object queries are conditioned on a language embedding.
Figure~\ref{fig:pipeline} illustrates this change.
This simple change has a few important implications: First, it allows the language embedding to control and focus queries to localize on the concepts at hand.
Second, it removes any contention between different object classes.
Each class present in the image uses the same amount of queries.
Third, it generalizes to unseen classes by leveraging semantic knowledge encoded in language embedding throughout the detection pipeline.
In the remainder of this section, we highlight the changes in the architecture and training objective for conditioning (\ours Phase 1), and self-training on image-level data for open-vocabulary detection (\ours Phase 2).

\myparagraph{Language-conditioned query selection.}
\ours conditions queries to a specific object category by modeling the objectness function as a similarity score between a region feature $x_{i,j}$ and a text representation of a category name $t(y)$ using their cosine similarity:
\begin{equation}
s_y(x_{i,j})=\frac{\langle x_{i,j}, t(y) \rangle}{\Vert x_{i,j}\Vert\Vert t(y) \Vert}, \;\; Q_y = \texttt{topk}_{x_{i,j}}(s_y(x_{i,j}))\label{eq:lvfunc} 
\end{equation}
The above objective avoids any inter-class calibration issue common in imbalanced data~\cite{seesaw,ecm,eql}. 
Queries do not compete, as \ours independently selects top-$k$ scoring regions $Q_y$ for each class $y$.
All queries proceed to the second stage in parallel.
A memory-efficient attention mechanism isolates interaction within each class.
After a series of decoding layers, each language-conditioned query predicts a \emph{single} scalar score, corresponding to the likelihood of the class $y$, and the associated box.
The overall architecture mirrors the two-stage deformable DETR~\cite{zhu2020deformable} with two modifications: a language-conditioned query, and a binary output classifier.

\myparagraph{Memory-efficient modeling.} \label{memory_efficient}
\ours uses $n = |Q_y|$ queries per class for $K$ classes.
Generally, $n$ is smaller than the total number of queries $|Q|$ of a standard DETR model.
However, since we produce $n$ queries per class, the total number of queries in \ours is much larger $|Q| \ll n K$.
A naive implementation of the DETR decoder is unable to cope with the $O(n^2K^2)$ memory requirements of the self-attention layers in the transformer decoder.
We thus modify the self-attention formulation to isolate it within each class, reducing the memory cost to $O(n^2K)$.
The actual implementation uses standard self-attention with a reshaping operation. See Figure~\ref{fig:pipeline} (right) for the illustration. 

\myparagraph{\ours Phase 1: Train to condition on given concepts.} 
Our goal is to design \ours to take a set of class names in an image (or a batch of images) and predict objects of the corresponding classes or backgrounds. 
For each class $y$, each conditioned query $q_y \in Q_y$ therefore only predicts a single \emph{presence score} for class $y$ and the box location. 
All predictions from the conditioned queries $P_y$ are matched with $G_y$, the subset of ground truth with class $y$: 
\begin{equation}
\sigma^*_y=\arg \min_{\sigma \in \mathfrak{S}_y} \ell (P_y, G_y|\sigma)\label{ours_objective}
\end{equation}
where $\mathfrak{S}_y$ is the set of possible matches between $P_y$ and $G_y$, and $\ell$ is the binary cross-entropy loss.
Unlike the original DETR objective in Eqn.~\ref{detr_objective}, Eqn.~\ref{ours_objective} matches within the conditioned class. 
It avoids \emph{inter-class competition} during training and simplifies the training objective. 
Instead, it learns to adapt its predictions to $y$; the set of conditioned query $q_y$ considers any objects other than $y$ as \emph{background}.

\myparagraph{Pseudo-labeling weakly-labeled data.}
\ours produces highly accurate predictions when conditioned on the exact categories of a scene, as shown in Section~\ref{sec:analyses}. 
This makes \ours a strong \emph{pseudo-labeler} for weakly-labeled data with either image tags or captions. 
We expand a large amount of such data with pseudo-bounding boxes of \ours Phase 1 and self-train altogether to scale up open-vocabulary object detection. 
Unlike other forms of weakly supervised learning such as knowledge distillation~\cite{vild} and online pseudo-labeling~\cite{detic,wu2023baron}, we simply generate labels for all images offline and jointly train over all pseudo-labeled data using the regular detection losses without any additional complication or slowdown. 
For each image and class $y$, \ours encodes the class' language feature and predicts a set of detections $P_y$. 
We simply choose the most confident prediction. 
Figure~\ref{fig:on_vs_off} shows our simple offline pseudo-labeling works better than online pseudo-labeling. 

\myparagraph{\ours Phase 2: Train for open-vocabulary detection.}
The advantage of \ours comes from adaptability to specified class names on a per-image basis.
However, in open-vocabulary detection, the set of test classes is neither known a priori nor available per image.
Hence, we convert \ours into a general-purpose detector to detect \textit{all objects}.
We condition \ours with ``$\texttt{an object}$'' as the text input, and inject the class information in the second-stage classifier. 
Figure~\ref{fig:pipeline} highlights this conversion.
Since \ours is trained to align image features to text embedding in both the first and second stages, this change only introduces inter-class calibration for multi-class object detection.
We train \ours with pseudo-labeled and human-labeled data as a standard supervised detection training, using the standard matching algorithm of DETR (Section~\ref{prelim_detr}). 
We do not introduce any additional hyper-parameter specifically for the weakly supervised learning~\cite{detic} or design choices~\cite{fvlm,zhong2022regionclip}, extra loss functions such as alignment loss~\cite{wu2023baron}, nor a large teacher model for knowledge distillation~\cite{vild,ov-detr}.
Generating pseudo-labels with \ours Phase 1 runs as fast as a regular detector and training Phase 2 is as easy as standard detection training on a supervised dataset. 
At a high level, \ours Phase 1 training objective optimizes for a strong pseudo-labeler instead of a multi-class detector, which differentiates \ours from prior works. 
Leveraging \ours Phase 1 to expand weakly-labeled data is the key contribution to scaling up the final open-vocabulary object detection.

\input{tables/ovddetr_all}
\input{tables/ovc2_full}
\input{tables/standard_all}

\section{Experiments}

We evaluate the effectiveness of \ours in two aspects: (1) pseudo-labeling quality of \ours Phase 1 (Section~\ref{sec:analyses}) and (2) benchmark evaluation (Section~\ref{sec:main_results}).
We consider three primary benchmarks to evaluate our final model (\ours Phase 2):
\emph{open-vocabulary LVIS}~\cite{vild}, \emph{standard LVIS}~\cite{gupta2019lvis}, and \emph{direct zero-shot evaluation} to LVIS, COCO~\cite{coco}, Object365~\cite{o365}, and OpenImages~\cite{oid}.

\subsection{Experimental Setup}
\label{sec:exp_setup}
\myparagraph{Datasets and benchmarks.}
We mainly evaluate our method on the LVIS dataset~\cite{gupta2019lvis}, a large-vocabulary instance segmentation and object detection dataset with 1203 naturally distributed object categories.
LVIS splits categories into \textit{frequent}, \textit{common}, and \textit{rare}.
For \emph{open-vocabulary LVIS}, we combine {frequent} and {common} categories into \emph{LVIS-base} and consider the {rare} categories as \emph{novel} concepts used for testing only~\cite{vild}. 
For \emph{standard LVIS}, we train and evaluate all classes. 
\emph{Direct zero-shot transfer} evaluates models trained on different detection data (e.g., Object365) and other weakly-labeled data without any prior knowledge about the target dataset such as the set of classes or object frequency. 
In this benchmark, we test \ours's generalization to different domains.
We evaluate \ours on LVIS, COCO~\cite{coco}, Object365~\cite{o365}, 
and OpenImages~\cite{oid} in a fully zero-shot manner.
All our models use the ImageNet-21K~\cite{ILSVRC15} dataset as weakly labeled data, which contains 14M of object-centric images annotated with a single class. 

\myparagraph{Evaluation metrics.} 
\label{sec:metric}
We evaluate \ours on AP$_{\text{novel/rare}}$, AP$_\text{c}$, AP$_\text{f}$, and mAP following the LVIS evaluation metric~\cite{gupta2019lvis}. 
\textit{We highlight the results in all three groups since we believe open-vocabulary detectors should not compensate for the performance of common/frequent classes for novel/rare classes.} 
We evaluate both AP$^\text{box}$ and AP$^\text{mask}$ for object detection and instance segmentation.
For \emph{zero-shot transfer} benchmark with COCO and Object365, we use AP, AP$_{50}$, and AP$_{75}$ following prior work~\cite{vild,detic}.
For OpenImages, we report AP$_{50}^\text{flat}$ on the expanded label space~\cite{detic,unidet}. 
For \emph{zero-shot transfer} to LVIS, we consider LVIS minival~\cite{mdetr} and standard LVIS v1.0 validation set and report AP$^\text{fixed}$~\cite{ap_fixed} following the prior works~\cite{mdetr,glip,groundingdino}. 
In addition, we pursue a more direct measurement of the generated pseudo-labeling quality.
Hence, we define \textbf{\emph{conditioned} mAP/AR} (c-mAP/AR) and compare it to baseline open-vocabulary detectors. 
c-mAP measures the detection performance in mAP \emph{when the detector is provided the set of ground truth classes in each image}. 
For example in Figure~\ref{fig:teaser}, both detectors use ``\textbf{\textcolor{mygreen}{cat}}'', ``\textbf{\textcolor{mytangoblue}{mentos}}'' and ``\textbf{\textcolor{myred}{cola}}'' as given.
This extra information is used to select scores to rank the final predictions (baselines), or directly condition the detector (\ours). 
We analyze the model's behavior and label quality in Section~\ref{sec:analyses} and Section~\ref{sec:additional_exp} in supplementary.

\input{tables/no_lvis_zeroshot}
\input{tables/cross_dataset}

\subsection{Models}
\label{sec:model}
\ours is based on two-stage Deformable DETR~\cite{zhu2020deformable}. 
As described in Section~\ref{sec:method}, the first-stage objectness function for query selection is replaced by a similarity score between the image feature and the CLIP text embedding of each class name. 
We train the detector with the improved DETR training recipe~\cite{zhang2022dino,ouyang2022nms}: \emph{look-forward-twice}, larger MLP hidden dimension, no dropout, \emph{etc}.
We consider four backbones: a ResNet-50~\cite{he2016deep}, Swin-B and L for all LVIS benchmarks, and Swin-T and L for the direct zero-shot transfer. 
Unless otherwise mentioned, all backbones are pretrained on the ImageNet-21K dataset~\cite{r50_21k}.
Next, we describe our key baseline models to directly compare to \ours. 

\myparagraph{{Baseline}.} We design a \emph{baseline} open-vocabulary detector to closely compare to \ours.
Inspired by Detic~\cite{detic}, \emph{baseline} replaces classification layers with the class embedding of the pretrained CLIP text encoder and is trained using Federated Loss~\cite{zhou2021probabilistic}.
\ours Phase 1 and \emph{baseline} are trained using human-labeled data (e.g., LVIS-base). 
All other settings (training dataset, number of training iterations, \emph{etc.}) are kept the same between \ours and \emph{baseline}.  

\myparagraph{{Baseline + self-train}.} 
Similar to \ours Phase 2, we self-train \emph{baseline} on weakly-labeled data.
For the self-training algorithm, we use online self-training with max-size loss from Detic~\cite{detic} as baseline comparison (\emph{baseline + self-train}) to \ours Phase 2. 
We tested max-size and max-score losses from Detic~\cite{detic} (online pseudo-labeling) as well as offline pseudo-labeling similar to \ours, and max-size loss consistently performed the best.

\myparagraph{\ours labels.} We train a two-stage detector for broader comparison: CenterNet2~\cite{zhou2021probabilistic}. 
Specifically, we use Detic's baseline model (``Detic-base''), a CenterNet2 trained on LVIS-base with CLIP embedding, and finetune on pseudo-labeled ImageNet-21K data using \ours Phase 1 of the same backbone size. 
We denote this as ``\ours labels''.

\myparagraph{Efficient modeling.}
For \ours Phase 1, we use $n=300$ queries per class. 
One memory and time bottleneck during \ours training is the first-stage loss computation. 
The original Deformable DETR computes Hungarian matching with all pixels to all objects in a class-agnostic manner, which is $\sum_{l\in L} H_l \cdot W_l$ predictions. 
To reduce the memory and time cost, we only consider the top $K=10,000$ confident pixels for each class $y$ during the first-stage matching and loss computation. 
Together with memory-efficient self-attention (Sec.~\ref{memory_efficient}), the training time and memory cost of \ours increases by less than $20 \%$ over the baselines (See Table~\ref{tab:runtime}). 

\myparagraph{Training details.} Following Detic~\cite{detic}, we train \ours Phase 1 and \emph{baseline} $4\times$ on LVIS-base, and further finetune for another $4\times$ on ImageNet-21K data with pseudo-labeling.
For training CenterNet2 with \ours labels, we combine pseudo-labels from different image resolutions for $h\in \{240, 280, 320, 360, 400\}$, where $h$ is the shorter side of the image. 
Note that this mimics the \emph{random image resizing} data augmentation during standard detection training. 
We use Detectron2~\cite{detectron2} based on PyTorch~\cite{pytorch} in all of the experiments. 
More details are in Section~\ref{sec:additional_details} of supplementary.

\subsection{Main Results}
\label{sec:main_results}
\myparagraph{Open-vocabulary LVIS.} 
Table~\ref{tab:ovddetr_all} compares \ours to  \emph{baseline} as well as state-of-the-art DETR-based open-vocabulary detectors; OV-DETR~\cite{ov-detr}, OWLv2~\cite{owlv2}, and \emph{baseline + self-train}.
For a fair comparison, we further finetune the official OV-DETR model checkpoint on ImageNet-21K for 4$\times$ schedule same as \ours Phase 2 and \emph{baseline + self-train}. 
For all backbone scales, we show consistent improvement over other methods. 
Notably, \emph{baseline + self-train} exhibits degradation in \emph{frequent} classes as a trade-off with improved \emph{novel} classes, which is commonly observed behavior in other open-vocabulary detection methods, too.
\ours improves all categories consistently, which highlights the quality of our pseudo-labels. 
In the last rows with the Swin-L backbone, we report the result of two concurrent works, DITO~\cite{dito} (Mask R-CNN-based) and OWLv2, to compare to the method that uses additional detection data (Object365) and billion-scale web data (DataComp-1B~\cite{datacomp}, WebLI~\cite{webli}). 
\ours demonstrates large improvement over the state-of-the-arts despite using orders of magnitude smaller training data and compute resources.
To further examine \ours's scalability, we test the pseudo-labeling capability of \ours on R-CNN-based detectors with \ours labels (Detic-base finetuned with our pseudo-labels).
In Table~\ref{tab:ovc2}, we compare {\ours labels to} a broad range of {literature} based on CenterNet2~\cite{zhou2021probabilistic} and Mask R-CNN~\cite{he2017mask}.
Table~\ref{tab:c2_r50} compares methods with ResNet-50 backbone and Table~\ref{tab:c2_sys} compares larger scale backbones for system-level comparison. 
In both tables, \ours clearly improve upon the state-of-the-art by large margins without additional complication in training and bells-and-whistles.

\begin{figure*}[t]
    \centering
    \begin{subfigure}{\columnwidth}
    \end{subfigure}
    \includegraphics[width=0.9\linewidth]{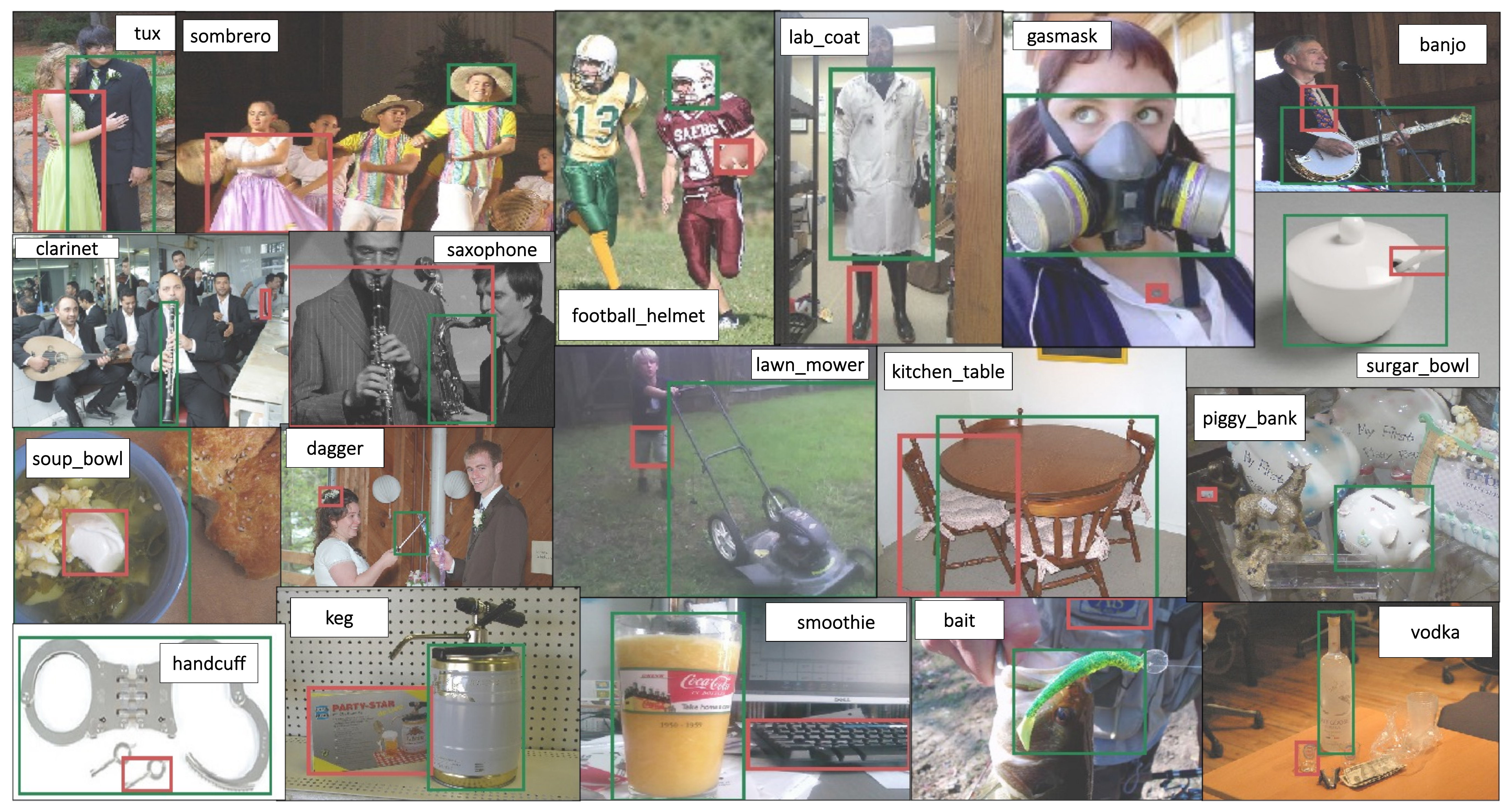}
    \captionsetup{width=\linewidth}
    \vspace{-2mm}
    \caption{
    \textbf{Examples of prediction on unseen categories.} Images from ImageNet-21K dataset. Boxes are the most confident prediction from \ours and \emph{\base}.
    \ours conditions on the ground truth label, and \textit{\base} selects the max-score box of the ground truth class.
    Images are all from unseen categories, which neither model was trained on. 
    \textbf{\textcolor{mygreen}{Green}}: \ours Phase 1 trained on LVIS-base. \textbf{\textcolor{myred}{Red}}: \emph{\base} trained on LVIS-base.
    Both models use a Deformable DETR detector with a ResNet-50 backbone. 
    More in Fig.~\ref{fig:random_labels1} and ~\ref{fig:random_labels2} of supplementary.
    }
    \vspace{-2mm}
    \label{fig:self_labels}
\end{figure*}

\begin{figure*}[t]
    \centering
    \includegraphics[width=\linewidth]{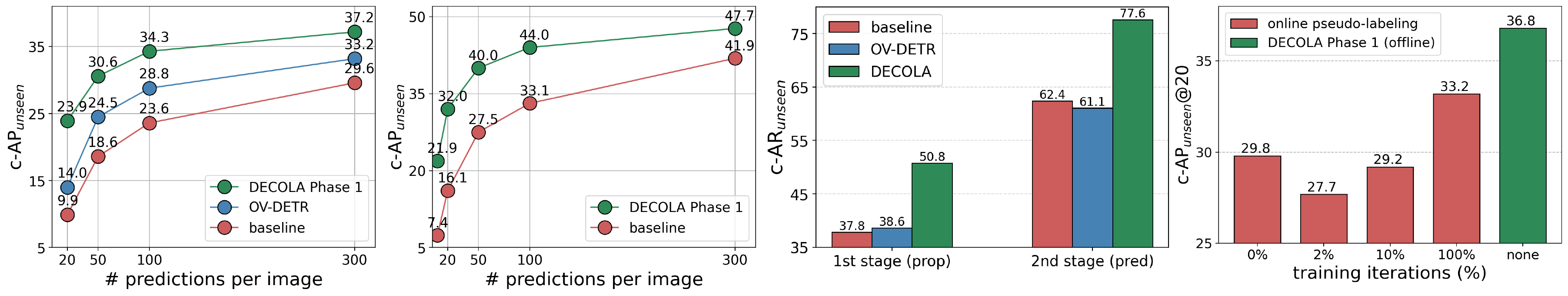}
    \begin{subfigure}{0.52\columnwidth}
        \captionsetup{
        justification=raggedright,
        singlelinecheck=true,
        width=\linewidth}
        \caption{{c-AP for ResNet-50 (IN-1K)}}
        \label{fig:cap_r50}
    \end{subfigure}
    \quad
    \begin{subfigure}{0.4\columnwidth}
        \captionsetup{width=\linewidth}       
        \caption{c-AP for Swin-B}
        \label{fig:cap_swin}
    \end{subfigure}
    \quad
    \quad
    \begin{subfigure}{0.5\columnwidth}
        \captionsetup{width=\linewidth}       
        \caption{c-AR for 1st and 2nd stage boxes}
        \label{fig:car}
    \end{subfigure}
    \begin{subfigure}{0.5\columnwidth}
        \captionsetup{width=\linewidth}       
        \caption{online vs offline: c-AP$@20$}
        \label{fig:on_vs_off}
    \end{subfigure}
    \captionsetup{width=\linewidth}
    \vspace{-2mm}
    \caption{\textbf{Analyzing \ours}. All plots show the \emph{conditioned} AP/AR for unseen classes. We compare \ours Phase 1 and baselines. 
    We highlight more detailed analyses about c-AP and c-AR in Section~\ref{sec:additional_exp}, Table~\ref{tab:oracle_ap_all} and~\ref{tab:oracle_map} of the supplementary materials.
    }
    \vspace{-2mm}
    \vspace{-3mm}
    \label{fig:analysis_all}
\end{figure*}

\input{tables/runtime}

\myparagraph{Standard LVIS.}
Tables~\ref{tab:st_all} evaluate \ours and \emph{baseline} on the standard LVIS benchmark, where all object categories are used to fully supervise the detectors. 
Similar to the open-vocabulary LVIS, we compare DETR architectures in Table~\ref{tab:st_detr} and R-CNN architectures in Table~\ref{tab:st_c2}. 
Table~\ref{tab:st_detr} shows that \ours remarkably improves \emph{baseline} by 9.5, 9.1, and 5.6 points on AP$^{\text{box}}_{\text{rare}}$, outperforming \emph{baseline + self-train} by 5.9, 5.4, and 6.2 AP$^{\text{box}}_{\text{rare}}$ for ResNet-50, Swin-B, and Swin-L backbones, respectively. 
Similarly in Table~\ref{tab:st_c2}, \ours labels further improves the baseline of Detic by 7.4 and 7.7 AP$^{\text{box}}_{\text{rare}}$, and outperforms Detic~\cite{detic} by 4.2 and 1.8 AP$^{\text{box}}_{\text{rare}}$ with ResNet-50 and Swin-B backbones, respectively.

\myparagraph{Direct zero-shot evaluation}.
For \emph{direct zero-shot evaluation}, we train \ours with Swin-T~\cite{liu2021Swin} and use Object365 data for Phase 1, and ImageNet-21K for Phase 2 (full dataset and classes). 
We compare to MDETR~\cite{mdetr}, GLIP~\cite{glip}, GroundingDINO~\cite{groundingdino}, and MQ-Det~\cite{mqdet} finetuned from GLIP and GroundingDINO. 
Table~\ref{tab:no_lvis_zs} shows the results. 
\ours outperforms the previous state-of-the-arts, by \textbf{12.0}/\textbf{17.1} AP$_\text{rare}$ and \textbf{3.0}/\textbf{9.4} mAP on LVIS minival and LVIS v1.0 val, respectively. 
It is noteworthy that all other methods use much richer detection labels from GoldG data~\cite{mdetr}, a collection of grounding data (box and text expression pairs) curated by MDETR.
Furthermore, other benchmark methods show highly imbalanced AP$_\text{rare}$ and AP$_\text{f}$ in both LVIS minival and LVIS v1.0 val (10-20 points gap). 
We hypothesize that the large collection of training data 
coincides with LVIS vocabulary, as all data follows a natural distribution of common objects. 
Also, \ours enjoys significantly faster run-time compared to all other models that undergo BERT encoding for grounding, which requires more than 50 forward passes per image in order to predict all LVIS categories. 
Similarly, our Swin-L model outperforms GroundingDINO and GLIP by \textbf{19.3} and \textbf{13.3} AP$_\text{rare}$, respectively, despite much smaller training data compared to FourODs~\cite{glip} and match to OWLv2 with 10-B private WebLI data~\cite{webli}. 
Table~\ref{tab:cross_dataset} further examines the generality of \ours on different domains. 
\ours Phase 2 with ResNet-50 outperforms all other competitive baselines by a large margin for both R-CNN and DETR architectures.

\subsection{Analyses}
\label{sec:analyses}
In this section, we analyze the model's behavior with \emph{conditioned} mAP/AR (c-AP/AR) (defined in Sec.~\ref{sec:metric}). 

\myparagraph{Pseudo-labeling quality.} Figure~\ref{fig:cap_r50} and~\ref{fig:cap_swin} compare \ours Phase 1, OV-DETR, and \emph{baseline} on c-AP for unseen classes. 
Compared to \emph{baseline} and OV-DETR, \ours Phase 1 generates much higher quality pseudo-labels, especially in low-shot regimes.
See examples in Figure~\ref{fig:self_labels}.

\myparagraph{Impact of conditioning.} In Figure~\ref{fig:car}, we compare c-AR of unseen classes. 
This measures the detector's ability to localize objects of interest when pseudo-labeling. 
We observe significant improvement in c-AR on both first-stage (proposals) and second-stage (predictions) due to our conditioning mechanism. 
This result demonstrates the key difference between \ours and other open-vocabulary detectors. 

\myparagraph{Pseudo-labeling algorithms.} Figure~\ref{fig:on_vs_off} shows the c-AP of \emph{baseline + self-train} and \ours Phase 1 for unseen classes with 20 predictions per-image. 
Each red bar indicates the percent of training iteration during self-training.
Online self-labeling suffers a sharp drop during the early iterations, and c-AP after full iterations still underperforms compared to \ours Phase 1.
\ours's simple approach of offline self-training is more stable and effective. 

\section{Conclusion}
In this paper, we explore a new open-vocabulary detection framework, \ours. 
It adjusts its inner workings to the concepts that the user asks to reason over by conditioning on a language embedding. 
Our detector generates high-quality pseudo-labels on weakly labeled data through the conditioning mechanism.
We finetune it with the pseudo-labels to build the state-of-the-art open-vocabulary detector. 

\myparagraph{Acknowledgement.} 
We thank Xingyi Zhou for his valuable feedback, and Yue Zhao, Jeffrey Ouyang-Zhang, and Nayeon Lee for fruitful discussions. 
This material is in part based upon work supported by the National Science Foundation under Grant No. IIS-1845485 and IIS-2006820.

{
    \small
    \bibliographystyle{ieeenat_fullname}
    \bibliography{main_cvpr2024}
}

\input{supp_cvpr2024}

\end{document}

%% file: tables/ovddetr_all.tex
\begin{table*}
\centering
\tablestyle{4pt}{1.05}
\begin{tabular}{ll|llll}
method  & data  & AP$^\text{box}_{\text{novel}}$  & AP$^\text{box}_{\text{c}}$  & AP$^\text{box}_{\text{f}}$ & mAP$^\text{box}$ \\
\shline
\textit{ResNet-50 (1K)} & & & & & \\
OV-DETR$^\dagger$~\cite{ov-detr} & LVIS-base, IN-21K & 18.0 & 24.8 & 31.8 & 26.4 \\
\base & LVIS-base & 10.2 & 30.9 & 38.0 & 30.1  \\
\base + self-train & LVIS-base, IN-21K & 19.2 & 31.7 & 37.1 & 31.7 \\
\rowcolor{mycell2}
\ours Phase 2 & LVIS-base, IN-21K & \goodnum{\textbf{23.8}}{4.6} & \goodnum{\textbf{34.4}}{2.7} & \goodnum{\textbf{38.3}}{1.2} & \goodnum{\textbf{34.1}}{2.4} \\
\midline
\textit{ResNet-50} & & & & & \\
\base & LVIS-base & 9.4 & 33.8 & 40.4 & 32.2 \\
\base + self-train & LVIS-base, IN-21K & 23.2 & 36.5 & 41.6 & 36.2 \\ 
\rowcolor{mycell2}
\ours Phase 2 & LVIS-base, IN-21K & \goodnum{\textbf{27.6}}{4.4} & \goodnum{\textbf{38.3}}{1.8} & \goodnum{\textbf{42.9}}{1.3} & \goodnum{\textbf{38.3}}{2.1} \\
\midline
\textit{Swin-B} & & & & & \\
\base & LVIS-base & 16.2 & 43.8 & 49.1 & 41.1 \\
\base + self-train & LVIS-base, IN-21K & 30.8 & 43.6 & 45.9 & 42.3 \\ 
\rowcolor{mycell2}
\ours Phase 2 & LVIS-base, IN-21K & \goodnum{\textbf{35.7}}{4.9} & \goodnum{\textbf{47.5}}{3.9} & \goodnum{\textbf{49.7}}{3.8} & \goodnum{\textbf{46.3}}{4.0} \\
\midline
\textit{Swin-L} & & & & & \\
DITO$^\star$~\cite{dito} & O365, LVIS-base, DataComp-1B & 45.8 & -& -& 44.2 \\ 
OWLv2$^\ddagger$~\cite{owlv2} & O365, LVIS-base, VG, WebLI & 45.9 & -& -& 50.4 \\
\base & O365, LVIS-base & 21.9 & 53.3 & 57.7 & 49.6 \\
\base + self-train & O365, LVIS-base, IN-21K & 36.5 & 53.5 & 56.5 & 51.8 \\ 
\rowcolor{mycell2}
\ours Phase 2 & O365, LVIS-base, IN-21K & \goodnum{\textbf{46.9}}{10.4} & \goodnum{\textbf{56.0}}{2.5} & \goodnum{\textbf{58.0}}{1.5} & \goodnum{\textbf{55.2}}{3.6} \\
\end{tabular}
\vspace{-2mm}
\caption{\textbf{Open-vocabulary LVIS} using DETR architectures. 
$\dagger$: we further finetune the official OV-DETR model trained on LVIS-base by self-training on ImageNet-21K data similar to ``baseline + self-train'' and \ours. 
$\ddagger$: uses CLIP L/14, which is comparable to Swin-L backbone. 
$\star$: is based on Mask R-CNN with ViT L/16.
We report the improvement between ``baseline + self-train'' to \ours in \textbf{\textcolor{mygreen}{green}}. 
Last 4 rows compare \ours to Swin-L or equivalent scale of models that use additional detection data (\emph{e.g.,} Objects365, VG~\cite{vg}) and billion-scale weakly-labeled data (DataComp-1B~\cite{datacomp}, WebLI~\cite{webli}) for training. 
}
\label{tab:ovddetr_all}
\end{table*}

%% file: tables/ovc2_full.tex
\begin{table*}[t]
\subfloat[
{Comparison with ResNet-50 backbone.}
\label{tab:c2_r50}
]{
\centering
\begin{minipage}{0.5\linewidth}{\begin{center}
\tablestyle{2pt}{1.05}
\begin{tabular}{ll|cc|cc}
method & framework & AP$^\text{box}_{\text{novel}}$ & mAP$^\text{box}$ & AP$^\text{mask}_{\text{novel}}$ & mAP$^\text{mask}$ \\
\shline
ViLD~\cite{vild} & Mask R-CNN & 16.7 & 27.8 & 16.6 & 25.5 \\
RegionCLIP~\cite{zhong2022regionclip} & Mask R-CNN & - & - & 17.1 & 22.5  \\
DetPro~\cite{du2022learning} & Mask R-CNN & 20.8 & 28.4 & 19.8 & 25.9 \\
PromptDet~\cite{promptdet} & Mask R-CNN & 21.4 & 25.3 & - & - \\
F-VLM~\cite{fvlm} & Mask R-CNN  & - & - & 18.6 & 24.2  \\
BARON~\cite{wu2023baron} & Mask R-CNN & 23.2 & 29.5 & 22.6 & 27.6 \\
OADP~\cite{oadp} & Mask R-CNN & 21.9 & 28.7 & 21.7 & 26.6 \\
EdaDet~\cite{edadet} & Mask R-CNN & - & - & 23.7 & 27.5 \\
VLDet~\cite{lin2023learning} & CenterNet2 & - & - & 21.7 & 30.1 \\
CORA$^+$~\cite{cora} & CenterNet2 & 28.1 & - & - & - \\
Rasheed et al.~\cite{rasheed} & CenterNet2 & - & - & 25.2 & 32.9 \\
Detic-base~\cite{detic} & CenterNet2 & 17.6 & 33.8 & 16.4 & 30.2 \\
Detic~\cite{detic}  & CenterNet2  & 26.7 & 36.3 & 24.6 & 32.4\\
\midline
\rowcolor{mycell1}
\ours labels & CenterNet2 & \textbf{29.5} & \textbf{37.7} & \textbf{27.0} & \textbf{33.7}\\
\end{tabular}
\end{center}}
\end{minipage}
}
\subfloat[
{System-level comparison.}
\label{tab:c2_sys}
]{
\centering
\begin{minipage}{0.5\linewidth}{\begin{center}
\tablestyle{4pt}{1.05}
\begin{tabular}{ll|cc|cc}
method & backbone & AP$^\text{box}_{\text{novel}}$ & mAP$^\text{box}$ & AP$^\text{mask}_{\text{novel}}$ & mAP$^\text{mask}$ \\
\shline
RegionCLIP~\cite{zhong2022regionclip} & R50$\times$4 & - & - &22.0 & 32.3 \\
CondHead~\cite{condhead} & R50$\times$4 & 24.1 & 33.7 & 24.4 & 32.0 \\
ViLD~\cite{vild} & EN-B7 & - & - & 26.3 & 29.3 \\
OWL-ViT~\cite{minderer2022simple} & ViT-L/14 & 25.6 & 34.7 & - & - \\
F-VLM~\cite{fvlm} & R50$\times$64 & - & - & 32.8 & 34.9 \\
VLDet~\cite{lin2023learning} & Swin-B & - & - & 26.3 & 38.1 \\
3Ways~\cite{3ways} & NFNet-F6 & 30.1 & 44.6 & - & - \\
RO-VIT~\cite{rovit} & ViT-L/16 & 32.1 & 34.0 & - & - \\
CFM-ViT~\cite{cfm} & ViT-L/16 & 35.6 & 38.5 & 33.9 & 36.6 \\
DITO~\cite{dito} & ViT-B/16 & 34.9 & 36.9 & 32.5 & 34.0 \\
CoDet~\cite{codet} & Swin-B & - & - & 29.4 & 39.2 \\
Detic-base~\cite{detic}  & Swin-B & 24.6 & 43.0 & 21.9 & 38.4 \\
Detic~\cite{detic}  & Swin-B & 36.6 & 45.7& 33.8 & 40.7 \\ 
\midline
\rowcolor{mycell1}
\ours labels  & Swin-B & \textbf{38.4} & \textbf{46.7} & \textbf{35.3} & \textbf{42.0} \\
\end{tabular}
\end{center}}
\end{minipage}
}
\vspace{-2mm}
\caption{\textbf{Open-vocabulary LVIS} using Mask R-CNN and CenterNet2 detectors. Methods in both tables use LVIS-base as the only human-labeled data for fair comparison. For system-level comparison (\textbf{right}), we include methods with non R-CNN architectures such as OWL-ViT~\cite{minderer2022simple}. The results show the impact of high-quality pseudo-labels generated by \ours Phase 1.   
} 
\vspace{-4mm}
\label{tab:ovc2}
\end{table*}

%% file: tables/standard_all.tex
\begin{table}[t]
\subfloat[
{Standard LVIS with DETR.}
\label{tab:st_detr}
]{
\centering
\begin{minipage}{\linewidth}{\begin{center}
\tablestyle{4pt}{1.05}
\begin{tabular}{ll|ll}
method  & data & AP$^\text{box}_{\text{rare}}$  &  mAP$^\text{box}$ \\
\shline
\textit{ResNet-50} & & & \\
\base & LVIS & 26.3 & 35.6 \\
\base + self-train & LVIS, IN-21K & 30.0 & 36.6 \\ 
\rowcolor{mycell2}
\ours Phase 2 & LVIS, IN-21K & \goodnum{\textbf{35.9}}{5.9} & \goodnum{\textbf{39.4}}{2.8}  \\
\midline 
\textit{Swin-B} & & & \\
\base & LVIS & 38.3 & 44.5 \\
\base + self-train & LVIS, IN-21K & 42.0 &  45.2 \\ 
\rowcolor{mycell2}
\ours Phase 2 & LVIS, IN-21K & \goodnum{\textbf{47.4}}{5.4} &  \goodnum{\textbf{48.3}}{3.1}  \\
\midline 
\textit{Swin-L} & & & \\
\base & O365, LVIS & 49.3 & 54.4 \\
\base + self-train & O365, LVIS, IN-21K & 48.7 & 53.4 \\ 
\rowcolor{mycell2}
\ours Phase 2 & O365, LVIS, IN-21K & \goodnum{\textbf{54.9}}{6.2} & \goodnum{\textbf{56.4}}{3.0}  \\
\end{tabular}
\end{center}}
\end{minipage}
}\\
\subfloat[
{Standard LVIS with CenterNet2.}
\label{tab:st_c2}
]{
\centering
\begin{minipage}{\linewidth}{\begin{center}
\tablestyle{5pt}{1.05}
\begin{tabular}{l|ll|ll}
method  & AP$^\text{box}_{\text{rare}}$  & mAP$^\text{box}$ & AP$^\text{mask}_{\text{rare}}$ & mAP$^\text{mask}$ \\
\shline
\textit{ResNet-50} & & & & \\
Detic-base~\cite{detic} &  28.2 & 35.3 & 25.6 & 31.4 \\
Detic~\cite{detic}  &  31.4 & 36.8 & 29.7 & 33.2 \\
\rowcolor{mycell2}
\ours labels &  \goodnum{\textbf{35.6}}{4.2} &  \goodnum{\textbf{38.6}}{1.8} & \goodnum{\textbf{32.1}}{2.4} & \goodnum{\textbf{34.4}}{1.2} \\
\midline 
\textit{Swin-B} & & & & \\
Detic-base~\cite{detic}  & 39.9 & 45.4 & 35.9 & 40.7 \\
Detic~\cite{detic}  & 45.8 & 46.9 & 41.7 & 41.7 \\
\rowcolor{mycell2}
\ours labels & \goodnum{\textbf{47.6}}{1.8} & \goodnum{\textbf{48.5}}{1.6} & \goodnum{\textbf{43.7}}{2.0} & \goodnum{\textbf{43.6}}{1.9} \\ 
\end{tabular}
\end{center}}
\end{minipage}
}
\vspace{-2mm}
\caption{\textbf{Standard LVIS benchmark}. \ours shows consistent improvement over different model scales and architectures.}
\label{tab:st_all}
\vspace{-6mm}
\end{table}

%% file: tables/no_lvis_zeroshot.tex
\begin{table*}[h]
\centering
\tablestyle{4pt}{1.05}
\begin{tabular}{ll|cccc|cccc}
 & & \multicolumn{4}{c}{LVIS minival} & \multicolumn{4}{c}{LVIS v1.0 val} \\
method  & data & AP$^\text{box}_{\text{rare}}$  & AP$^\text{box}_{\text{c}}$  & AP$^\text{box}_{\text{f}}$ & mAP$^\text{box}$ & AP$^\text{box}_{\text{rare}}$  & AP$^\text{box}_{\text{c}}$  & AP$^\text{box}_{\text{f}}$ & 
mAP$^\text{box}$ 
\\
\shline
\emph{Swin-T} & & & & && & & \\
MDETR$^\star$~\cite{mdetr} & \textcolor{mygrey}{LVIS}, GoldG, RefC & \color{mygrey} 20.9 & \color{mygrey} 24.9 & \color{mygrey} 24.3 & \color{mygrey} 24.2 & \color{mygrey} 7.4 & \color{mygrey} 22.7 & \color{mygrey} 25.0 & \color{mygrey} 22.5 \\ 
GLIP~\cite{glip} & O365, GoldG, Cap4M & 20.8 & 21.4 & 31.0 & 26.0 & 10.1 & 12.5 & 25.5 & 17.2 \\
GroundingDINO~\cite{groundingdino} & O365, GoldG, Cap4M & 18.1 & 23.3 & \textbf{32.7} & {27.4} &  - &- &- &-  \\
GLIPv2~\cite{glipv2} & O365, GoldG, Cap4M & - & - & - & 29.0 & - & - & - & - \\
MQ-GroundingDINO$^\dagger$~\cite{mqdet} & O365, GoldG, Cap4M, \textcolor{mygrey}{LVIS-5VQ} & \color{mygrey} 21.7 & \color{mygrey} 26.2 & \color{mygrey} 35.2 & \color{mygrey} 30.2 &\color{mygrey}  12.9 & \color{mygrey} 17.4 & \color{mygrey} 31.4 & \color{mygrey} 22.1 \\
MQ-GLIP$^\dagger$~\cite{mqdet} & O365, GoldG, Cap4M, \textcolor{mygrey}{LVIS-5VQ} & \color{mygrey}  21.0 & \color{mygrey}  27.5 &\color{mygrey}  34.6 & \color{mygrey} 30.4 & \color{mygrey} 15.4 &\color{mygrey}  18.4 & \color{mygrey} 30.4 &\color{mygrey}  22.6 \\
\rowcolor{mycell1}
\ours Phase 2 & O365, IN-21K$^\ddagger$ & {\textbf{32.8}}  & {\textbf{32.0}} & {31.8} & {\textbf{32.0}}  & \textbf{27.2} & \textbf{24.9} & \textbf{28.0} & \textbf{26.6} \\
\rowcolor{mycell1}
$\Delta$ & & \gooddelta{12.0} & \gooddelta{8.7} & - & \gooddelta{3.0} & \gooddelta{17.1} & \gooddelta{12.4} & \gooddelta{2.5} & \gooddelta{9.4} \\
\midline
\emph{Swin-L} & & & & & & & & \\
GLIP~\cite{glip} & FourODs, GoldG, Cap24M & 28.2 & 34.3 & \textbf{41.5} & 37.3 & 17.1 & 23.3 & \textbf{35.4} & 26.9 \\
GroundingDINO~\cite{groundingdino} & O365, OI, GoldG, Cap4M, COCO, RefC & 22.2 & 30.7 & 38.8 & 33.9 &  - &- &- &-  \\
MQ-GLIP$^\dagger$~\cite{mqdet} & FourODs, GoldG, Cap24M, \textcolor{mygrey}{LVIS-5VQ} & \color{mygrey}  34.5 & \color{mygrey}  41.2 &\color{mygrey}  46.9 & \color{mygrey} 43.4 & \color{mygrey} 26.9 &\color{mygrey}  32.0 & \color{mygrey} 41.3 &\color{mygrey}  34.7 \\
OWLv2~\cite{owlv2} & O365, VG, WebLI & 39.0 & - & - & \textbf{38.1} & \textbf{34.9} & - & - & \textbf{33.5} \\
\rowcolor{mycell1}
\ours Phase 2 & O365, OI, IN-21K$^\ddagger$ &\textbf{ 41.5} &\textbf{ 38.0} & 34.9 & 36.8 & 32.9 & \textbf{29.1} & 30.3 & 30.2 \\
\rowcolor{mycell1}
$\Delta$ & & \gooddelta{2.5} & \gooddelta{3.7} & - & - & - & \gooddelta{5.8} & - & - \\
\end{tabular}
\vspace{-2mm}
\caption{\textbf{Direct zero-shot transfer to LVIS.}
$\dagger$: methods use 5 per-class \emph{vision queries} of LVIS dataset (denoted as ``LVIS-5VQ''), which use images and annotations to extract instance-level features. 
$\star$: MDETR uses ResNet-101 backbone and trained fully-supervised on LVIS. 
Results that use LVIS data are in \colorbox{chosencolor}{gray}. 
$\ddagger$: Full ImageNet-21K.
No LVIS information is used in \ours.
}
\vspace{-6mm}
\label{tab:no_lvis_zs}
\end{table*}

%% file: tables/cross_dataset.tex
\begin{table}[t]
\centering
\tablestyle{4pt}{1.05}
\begin{tabular}{l|ccc|ccc|c}
 &\multicolumn{3}{c}{COCO} & \multicolumn{3}{c}{O365} & OI  \\
method  & AP & AP$_{50}$ & AP$_{75}$ & AP & AP$_{50}$  & AP$_{75}$  & AP$^{\text{flat}}_{50}$ \\
\shline
\emph{R-CNNs} & & & & & & &  \\
ViLD~\cite{vild}  & 36.6 & 55.6 & 39.8 & 11.8 & 18.2 & 12.6  & - \\
F-VLM~\cite{fvlm}  & 32.5 & 53.1 & 34.6 & 11.9 & 19.2 & 12.6 & -\\
DetPro~\cite{du2022learning}  & 34.9 & 53.8 & 37.4 & 12.1 & 18.8 & 12.9 & -\\
BARON~\cite{wu2023baron}  & 36.3 & 56.1 & 39.3 & 13.6 & 21.0 & 14.5 & - \\
Detic~\cite{detic}    & 39.1 & 56.3 & 42.2 & 14.2 & 20.7 & 15.2 & 42.9 \\
\midline
\emph{DETRs} & & & & & & & \\
OV-DETR~\cite{ov-detr}  & 38.1 & 58.4 & 41.1 & - & - & -  & -  \\
Detic~\cite{detic}    & 39.8 & 56.6 & 43.3 & 14.5 & 21.4 & 15.5 &41.6 \\
\midline
\rowcolor{mycell1}
\ours Phase 2  & \textbf{40.3} & \textbf{57.0} &\textbf{43.7} & \textbf{15.0} & \textbf{22.0} & \textbf{16.0} & \textbf{43.3}  \\
\end{tabular}
\vspace{-2mm}
\caption{\textbf{Cross-dataset generalization benchmark} on COCO, Object365, and OpenImages. All models use a ResNet-50 backbone and train on LVIS-base and weakly-labeled data. 
}
\label{tab:cross_dataset}
\vspace{-6mm}
\end{table}

%% file: tables/runtime.tex
\begin{table}[t]
\centering
\tablestyle{4pt}{1.05}
\begin{tabular}{l|cc|cc}
 & \multicolumn{2}{c}{train} & \multicolumn{2}{c}{test} \\
method &  time & mem. & time & mem. \\
\shline
\base & 44 h & 8.9 G & 0.07 s/img & 2.5 G\\
+ self-train & 45 h & 10.2 G & 0.07 s/img & 2.5 G \\ 
OV-DETR & 73 h & 22.0 G & {6.4} s/img & 3.4 G \\
\midline
\ours Phase 1 & 49 h & 12.6 G & - & - \\
\ours Phase 2 & 45 h & 8.9 G & 0.07 s/img & 2.8 G
\end{tabular}
\vspace{-2mm}
\caption{\textbf{Efficiency}. Training time is measured with 8 DGX V100 on ResNet-50, 2 images per-GPU. \texttt{float16} is used in both training and testing. OV-DETR uses the original \textit{optimized} inference. }
\vspace{-5mm}
\label{tab:runtime}
\end{table}

%% file: supp_cvpr2024.tex
\clearpage
\setcounter{page}{1}
\maketitlesupplementary

\begin{figure*}[t]
    \centering
    \includegraphics[width=0.9\linewidth]{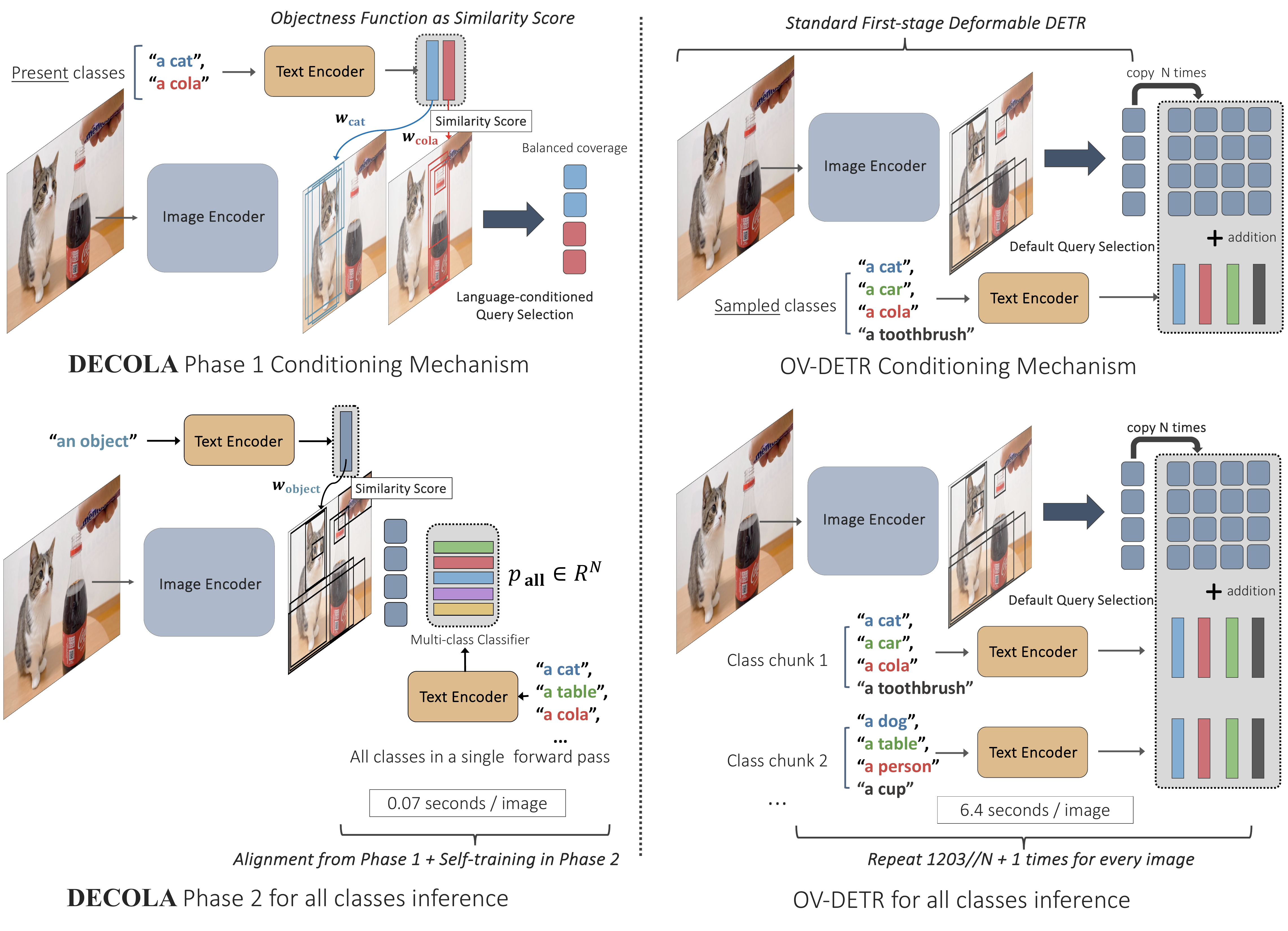}
    \caption{
    \textbf{Difference in conditioning mechanism and multi-class inference between \ours and OV-DETR.} \ours produces different \emph{proposals} for each present object class in image by modeling the objectness function with the similarity score between text embedding and proposal feature. 
    OV-DETR copies object queries from the first-stage DETR and add CLIP features. 
    Bottom row illustrates how \ours and OV-DETR performs multi-class detection.
    }
    \label{fig:diff}
\end{figure*}

\section{Comparing to OV-DETR~\cite{ov-detr}.} 

\myparagraph{Conditioning mechanism.}
Figure~\ref{fig:diff} illustrates the difference in conditioning mechanism and multi-class inference between \ours and OV-DETR. 
During training Phase 1, \ours learns to use text embedding of each present object class in order to locate proposals. 
\ours learns \textbf{dense language-vision alignment} by modeling the objecness function as the similarity score between text embedding and proposal features defined as equation~\ref{eq:lvfunc} in the Section~\ref{sec:method} of the main paper. 
\ours transforms equal number of proposals into query embedding to sufficiently cover all classes. 
On the other hand, OV-DETR trains with the same DETR object queries and add CLIP features of randomly sampled object classes. 
This difference results in a significant improvement in \emph{conditioned} AP and AR (+\textbf{9.1} c-AP$_{@20}$, +\textbf{16.5} c-AR), as shown in Figure~\ref{fig:cap_r50} and~\ref{fig:car} of the main paper. 

\myparagraph{Multi-class detection.}
\ours finetunes for multi-class object detection during Phase 2 whereas OV-DETR maintains the original conditioning mechanism for finetuning. 
Finetuning with multi-class detection objective is critical for the final detection task: Detector needs to \textbf{calibrate the multi-class scores} over the dataset-level vocabulary to maximize mAP.
OV-DETR trains with randomly sampled set of classes every iteration, which makes it unable to properly rank objects over all classes. 
This leads to a severe degradation in frequent classes as shown in Table~\ref{tab:ovddetr_all} of the main paper. 
Moreover, the conditioning mechanism of OV-DETR requires \textit{splitting} the text vocabulary over multiple chunks.
For LVIS dataset, OV-DETR needs about 40 forward passes for every image at inference, leading to a substantial difference in speed at run-time (0.07 vs 6.4 sec / img) as shown in Table~\ref{tab:runtime} of the main paper.
The final models, \ours Phase 2 and OV-DETR$^\dagger$ under identical training and architectural settings, exhibit large difference of \textbf{5.8} AP$_\text{novel}$ and \textbf{7.7} mAP, as shown in Table~\ref{tab:ovddetr_all} of the main paper.  

\myparagraph{Training setup.} 
Both models are trained on LVIS-base for $4\times$ (\ours Phase 1 and OV-DETR). 
OV-DETR undergoes extra $4\times$ with the original self-training using CLIP labeling~\cite{ov-detr}. 
This model is the same as the original OV-DETR reported in the original paper. 
We further finetune \ours and OV-DETR on ImageNet-21K for $4\times$ for fair comparison, which result \ours Phase 2 and OV-DETR$^\dagger$. 

\section{Experimental Details}
\label{sec:additional_details}
\myparagraph{Training configuration.} We closely follow~\cite{detic} to train \ours as well as \textit{baseline} for both Deformable DETR and CenterNet2 results.
Table~\ref{tab:training_configs} and~\ref{tab:model_configs} highlight important hyper-parameters in all experiments with Deformable DETR. 
For experiments with CenterNet2, we follow the same training and model configuration as Detic~\cite{detic}.
For all experiments, we used 8 V100 GPUs with 32G memory.
All models are trained on \texttt{float16} using Automatic Mixed Precision from PyTorch~\cite{pytorch}. 
With this computing environment, training \ours for Deformable DETR with ResNet-50 backbone takes about 50 hours and the baseline takes about 45 hours for $4\times$ training schedule. 
For ImageNet-21K pre-trained ResNet-50, we used the model from Ridnik \textit{et al.}~\cite{r50_21k} consistent with~\cite{detic}. 
Our codebase uses Detectron2~\cite{detectron2} based on PyTorch~\cite{pytorch}. 
For \emph{direct zero-shot transfer to LVIS} experiments, we use Swin-T and L~\cite{liu2021Swin} pretrained on ImageNet-21K. 
For both methods, we train Phase 1 on Object365 same number of iterations as GLIP~\cite{glip}. 
We finetune Phase 2 on the entire ImageNet-21K for Swin-T, and ImageNet-21K and OpenImages~\cite{oid} for the same number of iterations as Phase 1.
Please note that the model may continue to improve as training longer.
Swin-L model is trained with 2 nodes of 8 V100 machines, with 32 images per global batch. 
{All our experiments are conducted under academic-scale compute and open-sourced datasets. }

\input{tables/supp_hyperparams}

\input{tables/oracle_ap_rare}
\input{tables/oracle_map}

\input{tables/supp_results}

\input{tables/oracle_ap_diff_q}
\input{tables/oracle_map_diff_q}

\input{tables/backbone_ablation}

\section{Additional Experimental Results}
\label{sec:additional_exp}

\myparagraph{\textit{Conditioned} AP.} 
Table~\ref{tab:oracle_ap_all} and~\ref{tab:oracle_map} compare \textit{conditioned} mAP and AP$_{\text{novel}}$ of baseline and \ours Phase 1. 
We show AP with different per image detection limit, reported with $@k$. 
\textit{Conditioned} AP is defined in Section~\ref{sec:exp_setup}.
Results at low detection limit follows more closely to the labeling quality; pseudo-labels are sampled based on the confidence score and typically only save the top-1 prediction. 
\ours consistently improve baseline not only for novel classes but for overall. 
{This difference is the core reason for \ours's scalability by self-training. }

\myparagraph{Box-efficient detector.}
In this section, we highlight an interesting property of \ours. 
Object detectors for large-vocabulary dataset often tend to \textbf{over-shoot} predictions with a high number of boxes in order to increase recall for rare object classes. 
This behavior may be undesirable since lots of spammed boxes makes it difficult to interpret for downstream tasks. 
Therefore, Table~\ref{tab:oracle_ap_diff_q} and~\ref{tab:oracle_map_diff_q} report c-AP of baseline and \ours Phase 1 \textit{with a limited number of query (prediction) per class}. 
$n=1$ means the detector only gets to predict a single box for each class present in image. Please recall that c-AP provides a set of present classes during inference. 
We show that \ours show highly accurate predictions with low per-image detection limit. 

\myparagraph{Impact of different pre-training. } Table~\ref{tab:effect_backbone} shows how backbone pretraining impact the final result on \ours as well as the \textit{baseline}. 
In Table~\ref{tab:effect_backbone}, ImageNet-21K worked the best overall, but surprisingly there was no substantial difference in AP$_{\text{novel}}^{\text{box}}$ from Deformable DETR framework, contrary to the finding in~\cite{detic} with CenterNet2 detector. 
Here all models are trained on LVIS-base. 
Table~\ref{tab:pretrain_o365} shows that pretraining on Object365 substantially improve LVIS result. 
Since both Object365 and LVIS are large-scale detection datasets of natural objects, we expect some degree of semantic overlap between the datasets.

\myparagraph{Co-training. } 
\ours trains language-conditioning and multi-class prediction in two separate phases. 
Here, we explore if we can co-train both conditioning and multi-class prediction.
Specifically, we set a probability $p$ to train a detector by language-condition (conditioning the first-stage with class name) and multi-class (conditioning the first-stage with ``\texttt{an object}'') and using multi-class classifier with text embedding same as \textit{baseline}. 
Table~\ref{tab:cotrain1} reports the conditioned AP after training $4\times$ on LVIS-base with different $p$. 
We observe that c-AP is maximized with $p=0.0$, but mAP can match with the standard detection training with $p=0.5$. 
Table~\ref{tab:cotrain2} extends co-training to finetuning for Phase 2 on weakly labeled data. 
Here $p_1 \rightarrow p_2$ denotes the sampling probability of ``\texttt{a object}'' conditioning for LVIS-base ($p_1$) and LVIS-base and ImageNet-21K ($p_2$). 
We confirm that the quality of pseudo-labels is the most important for finetuning with weakly-labeled data.  

\myparagraph{Other ablations.} In Table~\ref{tab:boxloss}, we show that \ours label improves using box regression loss. 
Detic~\cite{detic} only trains for classification loss since max-size loss samples pseudo-label that does not localize object accurately. 
This improvement shows that \ours label provides a significant supervisory signal for localization as well as classification. 
In \ours Phase 1, each query is conditioned to an object class and predicts a \textit{single} score after decoding layers (``single''). 
Table~\ref{tab:second_stage_type} explores different second-stage formulation.
After the first stage, we ignore the conditioned classes and predict multi-class scores after decoding layers, denoted as ``multi''.

\begin{figure*}[h]
    \centering
    \includegraphics[width=\linewidth]{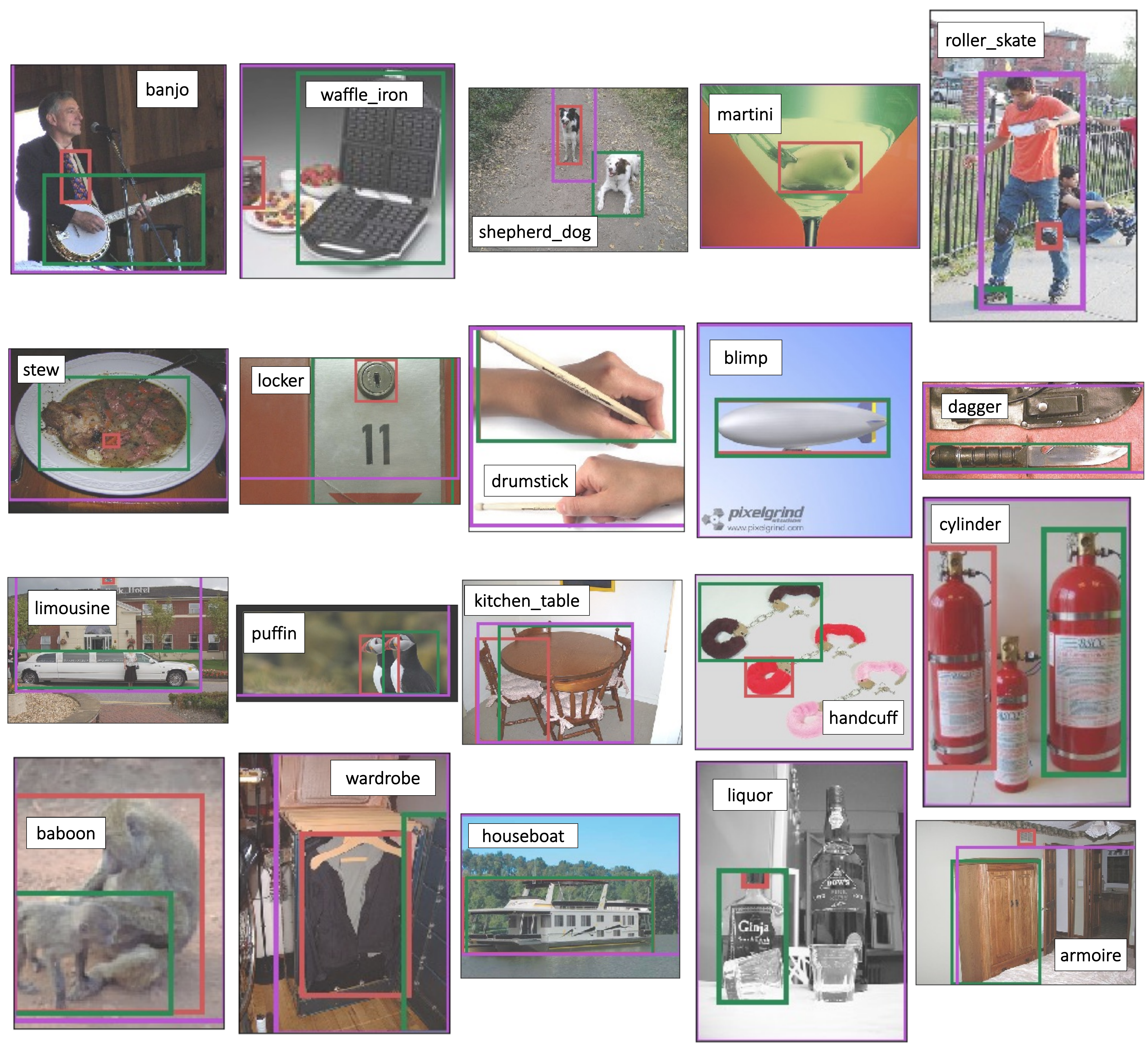}
    \caption{
    \textbf{Random samples of prediction on unseen categories.} 
    }
    \label{fig:random_labels1}
\end{figure*}
\begin{figure*}[h]
    \centering
    \includegraphics[width=\linewidth]{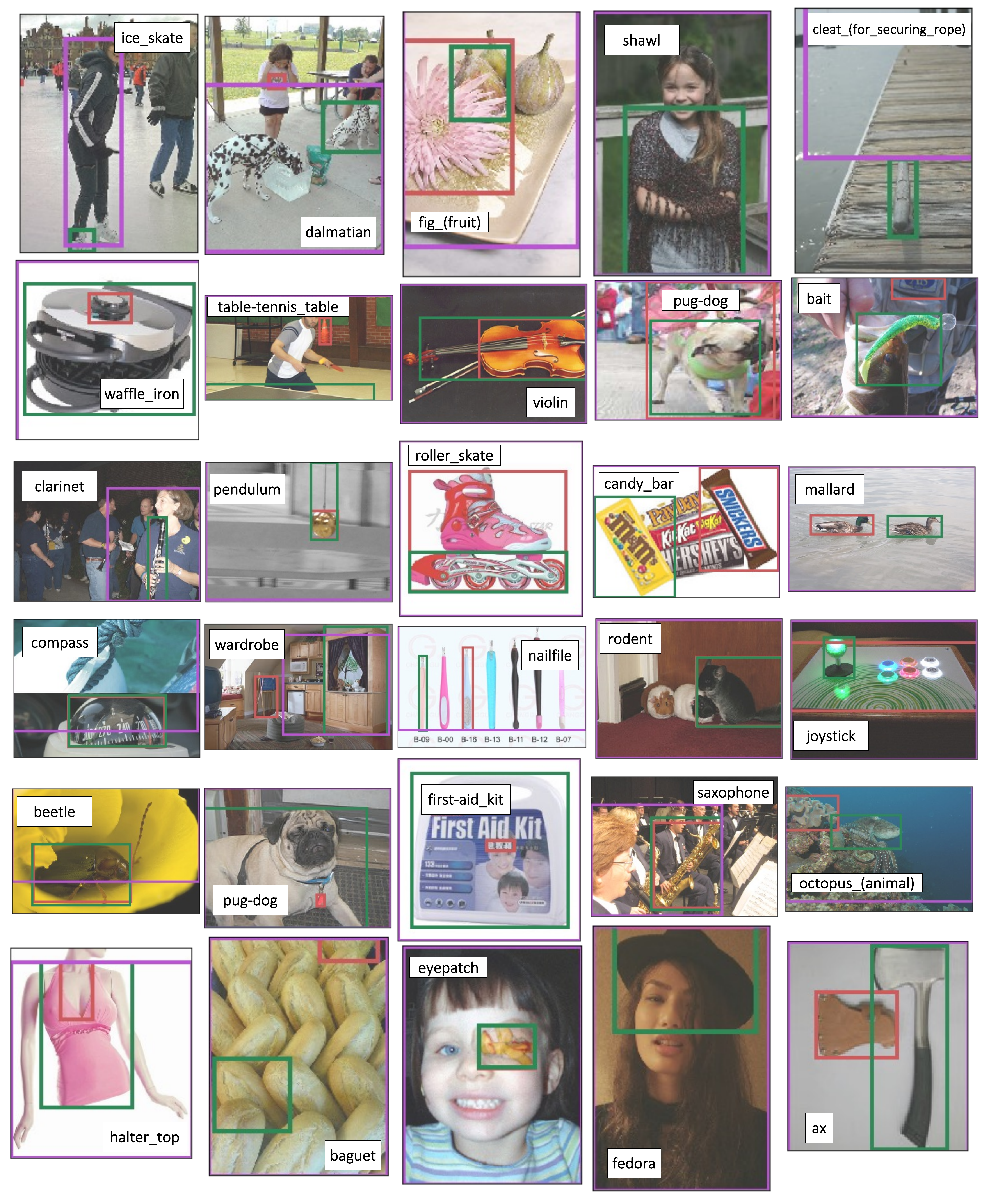}
    \caption{
    \textbf{Random samples of prediction on unseen categories.} 
    }
    \label{fig:random_labels2}
\end{figure*}

\section{Qualitative Results} 
We show more visualization.
Figure~\ref{fig:random_labels1} and~\ref{fig:random_labels2} show randomly sampled images and the pseudo-labels of \ours and baseline. 
Images are from the ImageNet-21K from \textit{unseen} categories, which none of the models are trained on. 
Boxes are the most confident prediction from \ours and \emph{\base} and maximum size box (\cite{detic}).
\textbf{\textcolor{mygreen}{Green}}: the most confident prediction (max-score) \ours trained on LVIS-base. 
\textbf{\textcolor{myred}{Red}}: the most confident prediction (max-score) \emph{\base} trained on LVIS-base.
\textbf{\textcolor{mypurple}{Purple}}: the largest box prediction (max-size, Detic loss~\cite{detic}) \emph{\base} trained on LVIS-base.
All models use a Deformable DETR detector with a ResNet-50 backbone.
We show randomly sampled images.

%% file: tables/supp_hyperparams.tex
\definecolor{chosencolor}{gray}{.8}
\newcommand{\chosen}[1]{\cellcolor{chosencolor}{#1}}

\begin{table*}[t]
\subfloat[
{Training configuration. The values inside parentheses are for LVIS and ImageNet-21K, respectively.}
\lbltbl{training_configs}
\label{tab:training_configs}
]{
\centering
\begin{minipage}{\linewidth}{\begin{center}
\tablestyle{14pt}{1.05}
\begin{tabular}{l|cc}
config & baseline training & baseline + self-train \\
\shline
\emph{shared configuration} \\ 
optimizer & AdamW~\cite{adamw} & AdamW~\cite{adamw} \\
optimizer momentum & $\beta_1,\beta_2=0.9,0.999$ & $\beta_1,\beta_2=0.9,0.999$ \\
weight decay & 0.0001 & 0.0001 \\ 
total iterations & 360000 & 360000 \\
base learning rate & 0.0002 & 0.0002 \\ 
learning rate schedule & step decay & step decay \\  
learning rate decay factor & 0.1 & 0.1 \\
learning rate decay step & 300000 & 300000 \\
gradient clip  value & 0.01 & 0.01 \\ 
gradient clip norm & 2.0 & 2.0  \\
\midline
\emph{different configuration} \\
batch size & 16 & (16, 64) \\
dataset ratio & n/a & $1 : 4$~\cite{detic} \\
image min-size range & (480, 800) & ((480, 800), (240, 400))~\cite{detic} \\
image max-size & 1333 & (1333, 667)~\cite{detic} \\
input augmentation & DETR-style~\cite{detr} & resize shortest edge~\cite{detic}  \\
input sampling & repeated factor sampling~\cite{gupta2019lvis} & (repeated factor~\cite{gupta2019lvis}, random) \\
\end{tabular}
\end{center}}
\end{minipage}
}\\
\subfloat[
{Model configuration.}
\lbltbl{model_configs}
\label{tab:model_configs}
]{
\centering
\begin{minipage}{\linewidth}{\begin{center}
\tablestyle{4pt}{1.05}
\begin{tabular}{l|ccc}
config & baseline & \ours Phase 1 & \ours Phase 2 \\
\shline
\emph{shared configuration} \\ 
\texttt{cls} weight & 2.0 & 2.0 & 2.0 \\
\texttt{giou} weight & 2.0 & 2.0 & 2.0 \\
\texttt{l1} weight & 5.0 & 5.0 & 5.0 \\
two-stage & \texttt{True} & \texttt{True} & \texttt{True} \\
box refinement & \texttt{True} & \texttt{True} & \texttt{True}\\
feed-forward dim. & 1024 & 1024 & 1024 \\
{look-forward-twice} & \texttt{True} & \texttt{True} & \texttt{True}\\
drop-out rate & 0.0 & 0.0 & 0.0 \\ 
\midline 
\emph{different configuration} \\ 
number of queries & 300 & 300 per class & 300 \\
classification loss type & federated loss~\cite{zhou2021probabilistic} & biniary cross-entropy & federated loss~\cite{zhou2021probabilistic} \\
1st-stage classifier type & learnable & ``\texttt{a [class name].}'' &  ``\texttt{an object.}'' \\
1st-stage classifier norm & n/a & L2 & L2 \\ 
1st-stage classifier temp. & n/a & 50 & 50 \\ 
1st-stage top-$k$ per class$^\dagger$ & n/a & 10000 & n/a \\
2nd-stage classifier type &         ``\texttt{a [class name].}'' &        ``\texttt{a [class name].}'' &         ``\texttt{a [class name].}'' \\
2nd-stage classifier norm  & L2 & L2 & L2 \\
2nd-stage classifier temp. & 50 & 50 & 50 \\
classifier $\#$ classes$^\ddagger$ & 1203 & 1 & 1203 \\
classifier bias init. value & $- \log(0.99/0.01)$ & $- \log(0.99/0.01)$ & $- \log(0.99/0.01)$ \\
\end{tabular}
\end{center}}
\end{minipage}
}
\caption{\textbf{Configurations.} Training and model details for experiments  with Deformable DETR.  $\dagger$ is the top-$k$ only for Hungarian matching and loss computation to reduce computation, as explained in the main paper. 
$\ddagger$ is for LVIS experiments. Here \ours for language-condition training has $\#$ classes as 1, since the second-stage with language-conditioned query is binary classification as opposed to multi-class classification in baseline and open-vocabulary detection.}
\label{tab:hyperparams}
\end{table*}

%% file: tables/oracle_ap_rare.tex
\begin{table*}[t]
\centering
\subfloat[
c-AP of unseen categories at different $k$. 
\lbltbl{oracle_ap}
\label{tab:oracle_ap_open_vocab}
]{
\begin{minipage}{\linewidth}{\begin{center}
\tablestyle{6pt}{1.05}
\begin{tabular}{ll|lllll}
model & data &
c-AP$_{\text{novel}}^{\text{box}}$$_{@10}$ &
c-AP$_{\text{novel}}^{\text{box}}$$_{@20}$ & 
c-AP$_{\text{novel}}^{\text{box}}$$_{@50}$  & 
c-AP$_{\text{novel}}^{\text{box}}$$_{@100}$ & 
c-AP$_{\text{novel}}^{\text{box}}$$_{@300}$ \\
\shline
\textit{ResNet-50} & & & & & \\
baseline & LVIS-base & 6.0 & 11.3 & 19.2 & 26.8 & 31.9 \\ 
\ours Phase 1   & LVIS-base & \goodnum{19.4}{13.4} & \goodnum{28.5}{17.2} & \goodnum{34.1}{14.9} & \goodnum{38.7}{11.9} & \goodnum{40.0}{8.1} \\ 
\midline 
\textit{Swin-B} & & & & & \\
baseline & LVIS-base & 7.4 & 16.1 & 27.5 & 33.1 & 41.9  \\ 
\ours Phase 1   & LVIS-base & \goodnum{21.9}{14.5} & \goodnum{32.0}{15.9} & \goodnum{40.0}{12.5} & \goodnum{44.0}{6.9} & \goodnum{47.7}{5.8} \\ 
\end{tabular}
\end{center}}\end{minipage}
}
\\
\subfloat[
c-AP of rare categories at different $k$. 
\lbltbl{oracle_ap}
\label{tab:oracle_ap_large_vocab}
]{
\begin{minipage}{\linewidth}{\begin{center}
\tablestyle{9pt}{1.05}
\begin{tabular}{ll|lllll}
model & data &
c-AP$_{\text{rare}}^{\text{box}}$$_{@10}$ &
c-AP$_{\text{rare}}^{\text{box}}$$_{@20}$ & 
c-AP$_{\text{rare}}^{\text{box}}$$_{@50}$  & 
c-AP$_{\text{rare}}^{\text{box}}$$_{@100}$ & 
c-AP$_{\text{rare}}^{\text{box}}$$_{@300}$ \\
\shline
\textit{ResNet-50} & & & & & \\
baseline & LVIS & 21.3 & 29.4 & 36.9 & 41.1 & 44.6  \\ 
\ours  Phase 1  & LVIS & \goodnum{26.6}{5.3} & \goodnum{39.1}{9.7} & \goodnum{45.2}{8.3} & \goodnum{47.1}{6.0} & \goodnum{48.8}{4.2}  \\ 
\midline 
\textit{Swin-B} & & & & & \\
baseline & LVIS & 30.1 & 38.2 & 45.5 & 49.3 & 53.2 \\ 
\ours Phase 1   & LVIS & \goodnum{33.5}{3.4} & \goodnum{43.9}{5.7} & \goodnum{51.4}{5.9} & \goodnum{53.8}{4.5} & \goodnum{55.8}{2.6} \\ 
\end{tabular}
\end{center}}\end{minipage}
}
\caption{\textbf{\textit{Conditioned}  AP$_\text{rare/novel}$} result of \ours Phase 1 and baseline pre-trained on LVIS-base (\textit{top}) and LVIS (\textit{bottom}). \textit{Conditioned} AP measures detection performance when the set of object categories present in each image is given. \base adapts its classification layer to the classes and \ours conditions itself to the classes, as described in Section~\ref{sec:method} of the main paper.   }
\label{tab:oracle_ap_all}
\end{table*}

%% file: tables/oracle_map.tex
\begin{table*}[h]
\centering
\subfloat[
c-mAP of all categories at different $k$. 
\label{tab:oracle_map_open_vocab}
]{
\begin{minipage}{\linewidth}{\begin{center}
\tablestyle{5pt}{1.05}
\begin{tabular}{ll|lllll}
model & data &
c-mAP$^{\text{box}}$$_{@10}$ &
c-mAP$^{\text{box}}$$_{@20}$ & 
c-mAP$^{\text{box}}$$_{@50}$  & 
c-mAP$^{\text{box}}$$_{@100}$ & 
c-mAP$^{\text{box}}$$_{@300}$ \\
\shline
\textit{ResNet-50} & & & & & \\
baseline & LVIS-base & 24.4 & 29.8 & 35.0 & 37.9 & 40.2 \\ 
\ours Phase 1   & LVIS-base & \goodnum{30.0}{5.6} & \goodnum{36.8}{7.0} & \goodnum{41.9}{6.9} & \goodnum{44.2}{6.3} & \goodnum{45.6}{5.4} \\ 
\midline 
\textit{Swin-B} & & & & & \\
baseline & LVIS-base & 29.6 & 36.9 & 43.4 & 46.0 & 48.8  \\ 
\ours Phase 1  & LVIS-base & \goodnum{33.5}{3.9} & \goodnum{41.3}{4.4} & \goodnum{47.4}{4.0} & \goodnum{49.7}{3.7} & \goodnum{51.5}{2.7} \\ 
\end{tabular}
\end{center}}\end{minipage}
}
\\
\subfloat[
c-mAP of all categories at different $k$. 
\label{tab:oracle_map_large_vocab}
]{
\begin{minipage}{\linewidth}{\begin{center}
\tablestyle{6.5pt}{1.05}
\begin{tabular}{ll|lllll}
model & data &
c-mAP$^{\text{box}}$$_{@10}$ &
c-mAP$^{\text{box}}$$_{@20}$ & 
c-mAP$^{\text{box}}$$_{@50}$  & 
c-mAP$^{\text{box}}$$_{@100}$ & 
c-mAP$^{\text{box}}$$_{@300}$ \\
\shline
\textit{ResNet-50} & & & & & \\
baseline & LVIS & 27.3 & 33.4 & 38.8 & 41.2 & 43.1  \\ 
\ours Phase 1   & LVIS & \goodnum{31.1}{3.8} & \goodnum{38.5}{5.1} & \goodnum{43.7}{4.9} & \goodnum{45.6}{4.4} & \goodnum{47.1}{4.0}  \\ 
\midline 
\textit{Swin-B} & & & & & \\
baseline & LVIS & 33.3 & 40.4 & 46.2 & 48.6 & 50.5 \\ 
\ours  Phase 1  & LVIS & \goodnum{35.7}{2.4} & \goodnum{43.6}{3.2} & \goodnum{49.4}{3.2} & \goodnum{51.6}{3.0} & \goodnum{53.2}{2.7} \\ 
\end{tabular}
\end{center}}\end{minipage}
}
\caption{\textbf{\textit{Conditioned}  mAP} result of \ours in phase 1 and baseline pre-trained on LVIS-base (\textit{top}) and LVIS (\textit{bottom}). \textit{Conditioned} mAP measures detection performance when the set of object categories present in each image is known. \base adapts its classification layer to the classes and \ours condition itself to the classes, as described in Section~\ref{sec:method} of the main paper.  }
\label{tab:oracle_map}
\end{table*}

%% file: tables/supp_results.tex
\begin{table*}[h]
\centering
\subfloat[
\textbf{Box regression loss} for weak data. 
\lbltbl{boxloss}
\label{tab:boxloss}
]{
\centering
\begin{minipage}{0.32 \linewidth}{\begin{center}
\tablestyle{4pt}{1.05}
\begin{tabular}{lcll}
model & reg. loss & AP$^\text{box}_{\text{novel}}$ & mAP$^\text{box}$ \\
\shline
\ours label & & 27.6 & 36.6 \\
\ours label & {\checkmark} & {29.5} & {37.7} \\
\end{tabular}
\end{center}}\end{minipage}
}
\subfloat[
\textbf{\ours Phase 2 vs. baseline + \ours label}. 
\label{tab:phase2_vs_label}
]{
\centering
\begin{minipage}{0.32 \linewidth}{\begin{center}
\tablestyle{4pt}{1.05}
\begin{tabular}{lll}
model & AP$^\text{box}_{\text{novel}}$ & mAP$^\text{box}$ \\
\shline
\base + \ours label & 25.1 & 36.9 \\
\ours Phase 2 & 27.6 & 38.3 \\
\end{tabular}
\end{center}}\end{minipage}
}
\subfloat[
\textbf{Second-stage type } for Phase 1 ($k=20$).
\label{tab:second_stage_type}
]{
\centering
\begin{minipage}{0.32\linewidth}{
\begin{center}
\tablestyle{12pt}{1.05}
\begin{tabular}{lcc}
type & c-AP$^\text{box}_\text{novel}$ & c-mAP$^\text{box}$  \\
\shline
multi & 14.2 & 20.7 \\
single & 28.5 & 40.0 \\
\end{tabular}
\end{center}
}
\end{minipage}
}
\\
\subfloat[
\textbf{Query types} ($k=20$). 
\label{tab:query_types}
]{
\centering
\begin{minipage}{0.32\linewidth}{\begin{center}
\tablestyle{14pt}{1.05}
\begin{tabular}{ccc}
type & c-AP$_{\text{novel}}^{\text{box}}$ & c-mAP$^{\text{box}}$ \\
\shline
base & 20.9 & 30.4 \\
text & 21.2 & 31.6 \\
image & 22.3 & 35.1 \\ 
\end{tabular}
\end{center}}\end{minipage}
}
\subfloat[
\textbf{Co-training}: Phase 1 ($k=20$). 
\label{tab:cotrain1}
]{
\centering
\begin{minipage}{0.32\linewidth}{\begin{center}
\tablestyle{14pt}{1.05}
\begin{tabular}{ccc}
$p$ & c-AP$_{\text{novel}}^{\text{box}}$ & mAP$^{\text{box}}$ \\
\shline
$1.0$ & 10.7 & 30.2 \\
$0.5$ & 19.1 & 30.2 \\
$0.0$ & 22.3 & n/a \\ 
\end{tabular}
\end{center}}\end{minipage}
}
\subfloat[
\textbf{Co-training}: Phase 2. 
\lbltbl{cotrain}
\label{tab:cotrain2}
]{
\centering
\begin{minipage}{0.32\linewidth}{\begin{center}
\tablestyle{3pt}{1.05}
\begin{tabular}{ccccc}
$p$ & AP$_{\text{novel}}^{\text{box}}$ & AP$_{\text{c}}^{\text{box}}$ & AP$_{\text{f}}^{\text{box}}$ & mAP$^{\text{box}}$ \\
\shline
$0.5 \rightarrow 0.5$ & 21.0 & 31.9 & 37.0 & 32.0 \\
$0.5 \rightarrow 1.0$ & 20.8 & 33.2 & 37.8 & 32.9 \\
$0.0 \rightarrow 1.0$ & 23.8 & 34.4 & 38.3 & 34.1 \\ 
\end{tabular}
\end{center}}\end{minipage}
}
\\
\caption{\textbf{Additional results.} open-vocabulary LVIS results for various ablation study. We used Deformable DETR with ResNet-50 for all models here. For all results with c-AP, we use Phase 1. $k$ represents the detection limit per image. Note that the bottom row tables (d), (e), (f) are trained with \ours and \textit{baseline} trained using ResNet-50 pretrained with ImageNet-1K. }
\lbltbl{ablations} 
\end{table*}

%% file: tables/oracle_ap_diff_q.tex
\begin{table*}[h]
\centering
\subfloat[
\textit{Conditioned} AP of unseen categories with different number of queries \textit{per class}. 
\lbltbl{oracle_ap}
\label{tab:oracle_ap_ov_diff_q}
]{
\begin{minipage}{\linewidth}{\begin{center}
\tablestyle{4pt}{1.05}
\begin{tabular}{ll|lllll}
model & data &
$n=1$ &
$n=2$ & 
$n=5$  & 
$n=10$ & 
$n=20$ \\
\shline
\textit{ResNet-50} & & & & & \\
baseline & LVIS-base & 14.7 & 22.4 & 27.6 & 30.9 & 32.2 \\ 
\ours  Phase 1  & LVIS-base & \goodnum{25.2}{10.5} & \goodnum{31.4}{9.0} & \goodnum{36.0}{8.4} & \goodnum{37.9}{7.0} & \goodnum{39.9}{7.7} \\ 
\midline 
\textit{Swin-B} & & & & & \\
baseline & LVIS-base & 17.8 & 26.0 & 33.7 & 37.6 & 40.9  \\ 
\ours  Phase 1  & LVIS-base & \goodnum{31.0}{13.2} & \goodnum{37.3}{11.3} & \goodnum{44.1}{10.4} & \goodnum{46.2}{8.6} & \goodnum{47.2}{6.3} \\ 
\end{tabular}
\end{center}}\end{minipage}
}
\\
\subfloat[
\textit{Conditioned} AP of rare categories with different number of queries \textit{per class}. 
\lbltbl{oracle_ap}
\label{tab:oracle_ap_lv_diff_q}
]{
\begin{minipage}{\linewidth}{\begin{center}
\tablestyle{6pt}{1.05}
\begin{tabular}{ll|lllll}
model & data &
$n=1$ &
$n=2$ & 
$n=5$  & 
$n=10$ & 
$n=20$ \\
\shline
\textit{ResNet-50} & & & & & \\
baseline & LVIS & 17.8 & 24.8 & 33.0 & 38.7 & 42.3  \\ 
\ours Phase 1   & LVIS & \goodnum{29.7}{11.9} & \goodnum{36.7}{11.9} & \goodnum{41.8}{8.8} & \goodnum{45.9}{7.2} & \goodnum{48.3}{6.0} \\ 
\midline 
\textit{Swin-B} & & & & & \\
baseline & LVIS & 20.7 & 29.9 & 42.4 & 48.4 & 51.6 \\ 
\ours Phase 1   & LVIS & \goodnum{34.5}{13.8} & \goodnum{42.3}{12.4} & \goodnum{49.0}{6.6} & \goodnum{50.8}{2.4} & \goodnum{52.7}{1.1} \\ 
\end{tabular}
\end{center}}\end{minipage}
}
\caption{\textbf{\ours is more box-efficient (c-AP$_\text{rare/novel}$).} We measure \textit{conditioned} AP of rare/unseen classes (c-AP$_\text{rare/novel}$) of \ours Phase 1 and baseline pre-trained on LVIS-base (\textit{top}) and LVIS (\textit{bottom}) with different \emph{per-class} number of query. \ours uses $n=|Q_y|$ language-conditioned queries for each class in image. Baseline uses $n\cdot |C_x|$ object queries where $C_x$ is the set of object classes in image $x$. Two models use the same total number of object queries.  }
\label{tab:oracle_ap_diff_q}
\end{table*}

%% file: tables/oracle_map_diff_q.tex
\begin{table*}[h]
\centering
\subfloat[
\textit{Conditioned} mAP of all categories with different number of queries \textit{per class}. 
\lbltbl{oracle_ap}
\label{tab:oracle_ap_ov_diff_q}
]{
\begin{minipage}{\linewidth}{\begin{center}
\tablestyle{5pt}{1.05}
\begin{tabular}{ll|lllll}
model & data &
$n=1$ &
$n=2$ & 
$n=5$  & 
$n=10$ & 
$n=20$ \\
\shline
\textit{ResNet-50} & & & & & \\
baseline & LVIS-base & 14.8 & 22.2 & 29.6 & 33.6 & 35.2 \\ 
\ours Phase 1   & LVIS-base & \goodnum{24.5}{9.7} & \goodnum{31.5}{9.3} & \goodnum{37.9}{8.3} & \goodnum{41.1}{7.5}  & \goodnum{43.4}{8.2} \\ 
\midline 
\textit{Swin-B} & & & & & \\
baseline & LVIS-base & 18.0 & 26.9 & 36.6 & 41.0 & 43.4  \\ 
\ours Phase 1 & LVIS-base & \goodnum{28.0}{10.0} & \goodnum{35.2}{8.3} & \goodnum{42.7}{6.1} & \goodnum{46.5}{5.5} & \goodnum{48.9}{5.5}  \\ 
\end{tabular}
\end{center}}\end{minipage}
}
\\
\subfloat[
\textit{Conditioned} mAP of all categories with different number of queries \textit{per class}. 
\lbltbl{oracle_ap}
\label{tab:oracle_ap_lv_diff_q}
]{
\begin{minipage}{\linewidth}{\begin{center}
\tablestyle{6.5pt}{1.05}
\begin{tabular}{ll|lllll}
model & data &
$n=1$ &
$n=2$ & 
$n=5$  & 
$n=10$ & 
$n=20$ \\
\shline
\textit{ResNet-50} & & & & & \\
baseline & LVIS & 15.0 & 22.6 & 31.1 & 35.7 & 38.3  \\ 
\ours Phase 1   & LVIS & \goodnum{25.5}{10.5} & \goodnum{32.2}{9.6} & \goodnum{38.9}{7.8} & \goodnum{42.6}{6.9} & \goodnum{44.8}{6.5} \\ 
\midline 
\textit{Swin-B} & & & & & \\
baseline & LVIS & 18.3 & 27.2 & 37.6 & 42.4 & 44.4 \\ 
\ours Phase 1   & LVIS & \goodnum{29.3}{11.0} & \goodnum{36.8}{9.6} & \goodnum{44.0}{6.4} & \goodnum{47.7}{5.3} & \goodnum{50.1}{5.7} \\ 
\end{tabular}
\end{center}}\end{minipage}
}
\caption{\textbf{\ours is more box-efficient (c-mAP).} We measure \textit{Conditioned} mAP of \ours Phase 1 and baseline pre-trained on LVIS-base (\textit{top}) and LVIS (\textit{bottom}) with different \emph{per-class} number of query. \ours uses $n=|Q_y|$ language-conditioned queries for each class in image. Baseline uses $n\cdot |C_x|$ object queries where $C_x$ is the set of object classes in image $x$. Two models use the same total number of object queries. \textit{Conditioned} mAP measures with $k=300$ per-image detection limit.  }
\label{tab:oracle_map_diff_q}
\end{table*}

%% file: tables/backbone_ablation.tex
\begin{table*}[t]
\centering
\subfloat[
Impact of the backbone pretrain.
\label{tab:pretrain_backbone}
]{
\centering
\begin{minipage}{\linewidth}{\begin{center}
\tablestyle{8pt}{1.05}
\begin{tabular}{ll|llll}
method & pretrain  & AP$^\text{box}_{\text{novel}}$  & AP$^\text{box}_{\text{c}}$  & AP$^\text{box}_{\text{f}}$ & mAP$^\text{box}$ \\
\shline
\multirow{3}{*}{\base} & IN-1K & 10.2 & 30.9 & 38.0 & 30.1  \\
& RegionCLIP~\cite{zhong2022regionclip} & 9.1 & 32.6 & 39.9 & 31.4 \\
&  IN-21K & \badnum{9.4}{0.8} & \goodnum{33.8}{2.9} & \goodnum{40.4}{2.4} & \goodnum{32.2}{2.1} \\
\midline 
\multirow{2}{*}{baseline + self-train } & IN-1K & 19.2 & 31.7 & 37.1 & 31.7 \\
&  IN-21K & \goodnum{23.2}{4.0} & \goodnum{36.5}{4.8} & \goodnum{41.6}{4.5} & \goodnum{36.2}{4.5} \\
\midline 
\multirow{2}{*}{\ours Phase 2} &IN-1K & 23.8 & 34.4 & 38.3 & 34.1 \\
& IN-21K & \goodnum{27.6}{3.8} & \goodnum{38.3}{3.9} & \goodnum{42.9}{4.6} & \goodnum{38.3}{4.2}  \\
\end{tabular}
\end{center}}\end{minipage}
}\\
\subfloat[
Impact of Object365~\cite{o365} pretrain on \ours Phase 1.
\label{tab:pretrain_o365}
]{
\centering
\begin{minipage}{\linewidth}{\begin{center}
\tablestyle{7.5pt}{1.05}
\begin{tabular}{lc|llll}
method & O365  & c-AP$_{\text{rare}}^{\text{box}}$$_{@300}$  & c-AP$_{\text{c}}^{\text{box}}$$_{@300}$  & c-AP$_{\text{f}}^{\text{box}}$$_{@300}$ & c-mAP$^{\text{box}}$$_{@300}$ \\
\shline
\multirow{2}{*}{\ours Phase 1} &  & 54.6 & 52.7 & 52.3 & 52.9  \\
&  \checkmark & \goodnum{62.0}{7.4} & \goodnum{62.0}{9.3} & \goodnum{61.6}{9.3} & \goodnum{61.8}{8.9} \\
\end{tabular}
\end{center}}\end{minipage}
}
\caption{ \textbf{Impact of different pretraining.} Evaluated on open-vocabulary LVIS  with ResNet-50 Deformable DETR (top) and large-vocabulary LVIS with Swinl-L Deformable DETR (bottom). We explore the impact of different pretraining on the final mAP (top) and conditioned AP (bottom).  }
\label{tab:effect_backbone}
\end{table*}